\documentclass{article}
\PassOptionsToPackage{numbers}{natbib}

 \usepackage[main, final]{neurips_2026}

\usepackage[utf8]{inputenc} %
\usepackage[T1]{fontenc}    %
\usepackage{hyperref}       %
\usepackage{url}            %
\usepackage{booktabs}       %
\usepackage{amsmath}
\usepackage{amsfonts}       %
\usepackage{nicefrac}       %
\usepackage{microtype}      %
\usepackage{xcolor}         %
\usepackage{enumitem}  
\usepackage{graphicx}

\usepackage{wrapfig}
\usepackage{subcaption}
\usepackage{multirow}
\usepackage{makecell}
\usepackage{longtable}
\usepackage{needspace}

\newcommand{\methodname}{RECIPE}

\title{RECIPE: Procedural Planning via Grounding in Instructional Video}

\author{%
\makebox[\textwidth][c]{%
\begin{tabular*}{\textwidth}{@{\extracolsep{\fill}}ccc@{}}
Luigi Seminara\textsuperscript{1,2} &
Antonino Furnari\textsuperscript{2} &
Lorenzo Torresani\textsuperscript{1} \\
\texttt{\small seminara.l@northeastern.edu} &
\texttt{\small antonino.furnari@unict.it} &
\texttt{\small l.torresani@northeastern.edu}
\end{tabular*}%
}\\[0.75em]
\makebox[\textwidth][c]{\small \textsuperscript{1}Khoury College of Computer Sciences, Northeastern University, Boston}\\
\makebox[\textwidth][c]{\small \textsuperscript{2}Department of Mathematics and Computer Science, University of Catania, Italy}\\\\
\url{https://farsightlab.github.io/RECIPE/}
}

\begin{document}

\maketitle

\begin{abstract}
Visual planning asks a model to generate the remaining steps of a procedure
in natural language given a partial video context and a goal. Progress on
this task is bottlenecked by annotation: clean labeled datasets are small,
domain-narrow, and encode a single execution trajectory per example, even
though many valid orderings often exist. Large-scale instructional video
corpora such as HowTo100M offer orders of magnitude more procedural content,
but supervised fine-tuning on pseudo-labels extracted from their noisy ASR
narrations fails: segmentation and alignment errors propagate into training,
and the resulting supervision is still single-trajectory. We identify a key
asymmetry. Extracting clean step labels from noisy video is hard, but
verifying whether a generated step sequence is temporally grounded in ASR
transcripts is comparatively cheap and scales to millions of videos via
precomputed text embeddings. We exploit this asymmetry in \methodname{},
which uses grounding quality as a reward signal for Group Relative Policy
Optimization (GRPO), turning the noisy corpus into a verification signal
rather than a labeling source. The framework applies uniformly to two input
configurations of the planner: a Socratic pipeline in which a frozen
vision-language model rewrites the video into a textual history fed to the planner, and a Video
configuration in which the planner consumes video tokens directly. It
applies to annotated supervision, weak supervision, and fully unsupervised
training, where histories are sampled directly from the noisy corpus.
We evaluate on seven procedural benchmarks using a reference-based LLM-as-judge
protocol that scores generated plans across six procedural-quality criteria.
\methodname{}-RL improves over the base checkpoint at every scale we test
(0.5B, 3B, 7B) and on every benchmark, with macro rubric score gains of
$+7$ to $+8$ points in-domain at every scale and up to $+16$ points zero-shot.
It substantially outperforms supervised fine-tuning on both annotated and
pseudo-labeled continuations and improves over the base checkpoints at
all three scales even when trained entirely on the noisy corpus, without
curated procedural datasets. Human evaluation indicates agreement between
the LLM judge and human preferences, with a system-level Spearman
correlation of $\rho=0.900$. A diversity analysis on COIN further shows
that \methodname{}-RL preserves the generation variety that supervised
fine-tuning collapses.
\end{abstract}

\section{Introduction}
\label{sec:intro}

Intelligent assistants embedded in wearable devices such as Ray-Ban-style smart glasses~\cite{somasundaram2023projectaria} promise to guide users through complex procedural tasks in real time, helping a novice repair an unfamiliar appliance, walking a home cook through a recipe as it unfolds, or supporting a hobbyist on a craft project~\cite{wang2023holoassist}. They do so by observing what the user is doing and speaking the next steps aloud in natural language. Similar capabilities are increasingly central to embodied agents that follow vision-language instructions to execute multi-step plans~\cite{kim2024openvla}.
In all of these settings, the system must perform \emph{visual planning}: given a video context capturing the procedure so far and a goal specified in natural language, generate a sequence of subsequent procedural steps that brings the user closer to the goal~\cite{patel2023pretrained}. While most benchmarks for this task evaluate predictions against fixed action taxonomies~\cite{wang2023pdpp,islam2024propose,zhao2022p3iv}, deployment in the wild calls for fluent, natural-language instructions whose content is directly conditioned on the video input: a closed taxonomy cannot exploit information extracted from the field of view, such as the specific tools, ingredients, and contextual cues visible to the user, that enables customization and finer-grained specification of the instructions delivered. This motivates both the methodological choices and the evaluation design we pursue in this work.

Despite considerable recent progress, procedural planning models remain bottlenecked by annotation scale. Clean labeled datasets such as CrossTask~\cite{zhukov2019crosstask} and COIN~\cite{tang2019coin} are expensive to produce and narrow in domain. Large-scale instructional video corpora such as HowTo100M~\cite{miech2019howto100m} offer orders of magnitude more procedural content, but their automatically transcribed narrations are noisy, asynchronous, and lack reliable step segmentation. A recent line of work relies on grounding-based pseudo-labeling~\cite{zare2024rap,zhao2022p3iv,wang2023pdpp}, where video-text grounding extracts step labels for supervised fine-tuning (SFT), but this inherits the grounding errors and pins the model to one trajectory per example.

Reinforcement learning (RL) has recently emerged as a powerful tool for improving language models on reasoning tasks~\cite{shao2024deepseekmath}, but its successes have hinged on the availability of \emph{deterministic, verifiable rewards}: proof correctness, code execution, exact-match answers. Procedural planning offers no such crisp verifier, since a plan's validity is multidimensional, context-dependent, and admits many correct realizations, a structure that single-trajectory supervision is poorly equipped to exploit. Recent work responds by training learned \emph{critic models} to score plans for cost-based search~\cite{chen2025planning}, but the critic itself becomes another component requiring its own supervision pipeline.

\begin{figure*}[t]
    \centering
    \includegraphics[width=\textwidth]{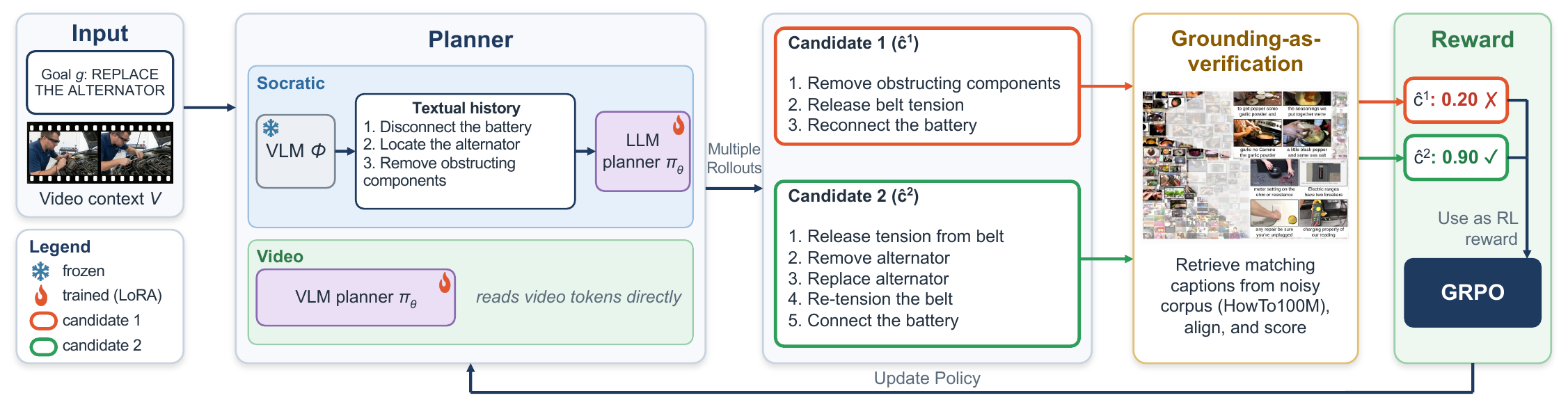}
    \caption{\textbf{Overview of \methodname{}.}
    The Socratic configuration uses a frozen VLM to convert video
    into textual history, while the Video configuration directly
    consumes video tokens. Both generate candidate plans scored
    against noisy instructional narrations from HowTo100M.
    The resulting rewards train the planner via GRPO.} %

    \label{fig:teaser}
\end{figure*}

Our key observation is that the same primitive, video-text grounding, that has been used to \emph{extract} labels from noisy video can also be used to \emph{verify} generated plans, and that verification is fundamentally easier and cheaper than label extraction. We exploit this asymmetry by reframing the role of the noisy corpus: rather than mining it for labels or distilling it into a critic model, we use the data \emph{itself} as the verifier. The history is used to localize the relevant region of the corpus, and the reward measures how well the generated completion aligns with that region, a density-estimation principle in which the data implicitly serves as the model of plan validity. Because many different orderings can each score well, the reward lets the model learn from several valid plans at once instead of being pushed toward a single annotator's choice, preserving the natural variety that SFT loses (\S\ref{sec:exp-diversity}).

Concretely, our framework, \methodname{} (Reward from Empirical Corpus of Instructional Procedure Examples; Figure~\ref{fig:teaser}), uses grounding quality as a reward signal for Group Relative Policy Optimization (GRPO), turning a noisy instructional video corpus into a scalable verifier of plan plausibility. We instantiate \methodname{} in two input configurations (Socratic and Video, \S\ref{sec:inputs}) and consider three training schemes (SFT only, RL only, SFT$\to$RL).

In summary, our contributions are:
\begin{enumerate}[leftmargin=*, topsep=0.2em, itemsep=0.3em, parsep=0pt, partopsep=0pt]
    \item \textbf{Grounding-as-verification for procedural planning.} We turn a noisy, large-scale instructional video corpus into a verifier for generated plans rather than a source of labels, with no pseudo-label extraction and no separately trained critic. A history baseline credits the policy only for what the continuation adds beyond the partial plan. The reward is optimized with GRPO and applies uniformly whether the planner reads textual histories (Socratic) or raw video tokens (Video).
    \item \textbf{Empirical study.} \methodname{}-RL improves over the base checkpoint at every scale (0.5B, 3B, 7B) and on all seven in-domain and zero-shot benchmarks, with macro rubric score gains of $+7$ to $+8$ points in-domain at every scale and up to $+16$ points zero-shot. It outperforms supervised fine-tuning on both annotated continuations and HowTo100M pseudo-labels, and transfers to a Video planner that consumes video tokens. Under fully unsupervised training, \methodname{}-RL improves over Base at all three scales without curated procedural training datasets. Human evaluation shows a system-level Spearman correlation of $\rho=0.900$ between judge and human rankings. On the COIN test split, \methodname{}-RL preserves the generation variety that supervised fine-tuning collapses.    
    \item \textbf{An open-ended evaluation protocol.} We aggregate seven procedural-video datasets and score generated plans with a six-criterion large language model (LLM)-as-judge rubric, supporting free-form evaluation beyond fixed-taxonomy benchmarks. We will host an online evaluation server alongside the released code, aiming to establish a reference benchmark for open-vocabulary procedural planning.
\end{enumerate}

\section{Related Work}
\label{sec:related_work}

\vspace{-1.0em}
\paragraph{Visual and Language-Based Planning.}
Early procedure planning models use sequence generators (diffusion: PDPP~\cite{wang2023pdpp}, KEPP~\cite{nagasinghe2024not}; transformers: SCHEMA~\cite{niu2024schema}) to learn procedural knowledge implicitly, inject explicit knowledge (ViterbiPlanNet~\cite{Seminara2026ViterbiPlanNet}), or use auxiliary tasks to mitigate annotation scarcity~\cite{Zhang_2026_WACV}. More recent LLM-based visual planners treat forecasting as sequence modeling from partial video~\cite{patel2023pretrained}, integrate retrieval via breadth-first search~\cite{islam2024propose}, or enforce consistency through world models~\cite{chen2025planning}. Concurrent open-vocabulary efforts include PlanLLM~\cite{yang2025planllm}, which pairs a trainable LLM with closed-set decoding from start and goal frames, and OEPP~\cite{wu2024oepp}, which extends procedure planning to unseen events. These methods either require domain-specific architectures or constrain outputs to a closed taxonomy. \methodname{} produces open-ended textual plans directly from visual contexts, with no domain-specific architecture or post-processing.

\vspace{-1.0em}
\paragraph{Step grounding and pseudo-labeling from narrations.}
A line of work mines procedural supervision by aligning narrations or wikiHow articles to instructional videos. MIL-NCE~\cite{miech2020milnce} learns video-text representations from noisy HowTo100M narrations; Drop-DTW~\cite{dvornik2021dropdtw} and StepFormer~\cite{dvornik2023stepformer} localize steps via sequence alignment; Mavroudi et al.~\cite{mavroudi2023grounding} ground wikiHow articles in HowTo100M with iterative pseudo-labels; Zhong et al.~\cite{zhong2023procedureaware} learn procedure-aware representations via a diffusion model over step orderings; and RAP~\cite{zare2024rap} expands training via retrieval-augmented pseudo-labels. \methodname{} reuses the alignment primitive as a verifier of generated plans rather than a labeler, so segmentation/alignment errors are absorbed by a soft reward instead of committed to hard labels.

\vspace{-1.0em}
\paragraph{Benchmarks for Procedure Planning.}
Existing procedure planning benchmarks rely on closed action taxonomies, e.g., CrossTask~\cite{zhukov2019crosstask}, COIN~\cite{tang2019coin}, NIV~\cite{alayrac2016unsupervised}; newer egocentric benchmarks like EgoPlan-Bench~\cite{qiu2024egoplan,chen2026egoplan} evaluate vision-language models (VLMs) but still depend on closed-vocabulary or multiple-choice formats. To match natural-language deployment in assistive and wearable systems, we introduce an open-ended planning benchmark evaluated via a reference-based LLM-as-judge protocol.

\vspace{-1.0em}
\paragraph{RL with verifiable rewards.}
The Reinforcement Learning with Verifiable Rewards (RLVR) paradigm~\cite{lambert2024tulu3,shao2024deepseekmath,guo2025deepseekr1} has driven gains on math and code by combining critic-free GRPO with rule-based correctness checks. RL has also been applied to LLM planning with tool-use rewards~\cite{li2025rltr} and offline goal-conditioned strategies~\cite{hong2025offlinegcrl}, and to video LLMs with QA or temporal-grounding rewards~\cite{feng2025videor1}. All rely on a deterministic verifier. \methodname{} operates without one: grounding scores against a noisy video corpus serve as a soft, trajectory-level reward, turning the corpus into a scalable verifier without pseudo-labels or a separately trained critic. \methodname{} also relates to reinforcement learning from human feedback (RLHF)~\cite{kirk2023understanding}, with the corpus replacing human preferences.

\section{Technical Approach}
\label{sec:techapproach}

\subsection{Problem setup}
\label{sec:setup}

We address the visual planning task in the spirit of~\cite{patel2023pretrained}:
given a partial video context capturing the procedure executed so far and a
goal $g$ specified in natural language, the model must produce a sequence of
remaining steps in natural language that, when executed, achieves $g$.
Unlike~\cite{patel2023pretrained}, we do not constrain predicted
steps to a closed taxonomy, allowing instead free-form natural-language
outputs that can be tailored to the specific tools, ingredients, and
contextual cues visible in the video. 
\vspace{-1.0em}
\paragraph{Annotated datasets.}
For training and evaluation we use a small collection of \emph{annotated}
procedural datasets (e.g., CrossTask~\cite{zhukov2019crosstask},
COIN~\cite{tang2019coin}, CaptainCook4D~\cite{CaptainCook4D}). Each example
takes the form $(V, g, \mathbf{y}^*)$, where $V$ is a video of the full
procedure and $\mathbf{y}^* = (y^*_1, \ldots, y^*_K)$ is a sequence of
time-stamped natural-language step descriptions covering it. 
Splitting $V$ and
$\mathbf{y}^*$ at a cut point $1 \le t < K$ yields a \emph{visual history}
$h_V = \{V_\tau : \tau < \text{time}(y^*_t)\}$ containing all frames up to the
time-stamp of step $y^*_t$, a \emph{textual history} $h = (y^*_1, \ldots, y^*_t)$,
and a \emph{continuation} $c = (y^*_{t+1}, \ldots, y^*_K)$. 
\vspace{-1.0em}
\paragraph{Input configurations.}
\label{sec:inputs}
We instantiate the policy $\pi_\theta$ as an autoregressive (vision-)language
model, in two configurations that differ in how $h_V$ is consumed. In the
\emph{Socratic configuration}, a frozen vision-language model $\Phi$ extracts
a textual summary of $h_V$ as a sequence of history steps
$\hat{h} = \Phi(h_V)$, which is then passed as text to a language model
$\pi_\theta(c \mid \hat{h}, g)$. This decouples perception from planning and
allows reuse of strong off-the-shelf VLMs and text-only LLMs without
architectural modification. In the \emph{Video configuration}, a
vision-language model $\pi_\theta(c \mid h_V, g)$ consumes video tokens
directly. In both configurations, $\pi_\theta$ is adapted with low-rank adaptation (LoRA)~\cite{hu2022lowrank} applied
to the query and value projections of the language decoder. The Video
configuration additionally fully fine-tunes a vision-to-text projector while
keeping the visual encoder frozen.
\vspace{-1.0em}
\paragraph{Supervision regimes.}
At test time the policy never has access to a clean textual history: it sees
only $h_V$ in the Video configuration, or its VLM-derived summary $\Phi(h_V)$
in the Socratic configuration. We consider three training regimes that differ
in their dependence on human annotations and curated datasets.
Under \emph{annotated supervision}, training videos are drawn from annotated
procedural datasets and $(h, c)$ is taken directly from $\mathbf{y}^*$ at a
sampled cut point, with multiple cut points per video used as data
augmentation. Under \emph{weak supervision}, the training videos come from
the same datasets, but the human $\mathbf{y}^*$ is discarded: we instead
apply $\Phi$ to $V$ to extract a pseudo-trace
$\hat{\mathbf{y}} = \Phi(V)$ and sample $(h, c)$ from it.
This matches the test-time input but inherits the VLM's errors.
Under \emph{fully unsupervised training}, we remove the curated datasets
entirely and sample training histories directly from HowToCaption, using
the same corpus to compute the grounding reward. Table~\ref{tab:configurations}
summarizes the input configurations, supervision regimes, and training schemes.

\subsection{Grounding-as-verification reward}
\label{sec:reward}

\paragraph{Verification rather than labeling.}
The reward in \methodname{} is computed against a large, noisy corpus of
instructional video. We use only its text narrations, not the video frames
themselves, and we use them only as a scoring target, not as a source of
labels for the policy. Concretely, we use the HowToCaption
release~\cite{shvetsova2024howtocaption}, which reprocesses
HowTo100M~\cite{miech2019howto100m} automatic speech recognition (ASR) transcripts into caption-like
descriptions, yielding 25M video-text pairs from 1.2M instructional videos
with no human annotation. These narrations form a set
$\mathcal{N} = \{n^{(1)}, n^{(2)}, \ldots\}$ in which each $n^{(j)}$ is a
sequence of time-stamped segments. The corpus is orders of magnitude larger
than the annotated datasets of \S\ref{sec:setup} but remains noisy, weakly
aligned with the visual content, and lacks reliable step segmentation, which
makes it a poor source of clean labels. The central design principle of our
reward is that this noisy $\mathcal{N}$ should serve as a \emph{verifier}
rather than a label source, with alignment performed entirely in a precomputed sentence-embedding space, making it tractable at the scale of $\mathcal{N}$.

\vspace{-1.0em}
\paragraph{Training-time alignment with the reference history.}
The reward against $\mathcal{N}$ is computed for a reference textual history $h$
that plays two roles: (i) anchoring retrieval in $\mathcal{N}$ to narrations
that depict similar partial executions, and (ii) defining a baseline against
which the generated continuation $\hat{c}$ is scored (see \emph{History-baseline
correction} below). 
The source of $h$ follows the supervision regime (\S\ref{sec:setup}, Table~\ref{tab:configurations}): under annotated supervision $h$ is the clean prefix $(y^*_1, \ldots, y^*_t)$, under weak supervision $h := \Phi(h_V)$, and under fully unsupervised training $h$ is sampled from HowToCaption. Policy conditioning is independent of $h$: the Video policy reads $h_V$ directly, while the Socratic policy reads either $h$ or $\hat{h} = \Phi(h_V)$ to match the test-time input distribution.
Concretely, given a training tuple $(h_V, g, h, c)$ and a generated completion
$\hat{c} \sim \pi_\theta$, we form two text sequences
\begin{equation}
    s_{\text{full}} = h \oplus \hat{c}, \qquad s_{\text{hist}} = h,
    \label{eq:sfull-shist}
\end{equation}
where $\oplus$ denotes concatenation at the step level. Both sequences are scored
against $\mathcal{N}$ using the alignment procedure described next.

\vspace{-1.0em}
\paragraph{Two-stage alignment against the corpus.}
For each candidate sequence $s = (s_1, \ldots, s_M)$, we compute a grounding score
$A(s, \mathcal{N})$ that measures how well $s$ matches the most procedurally similar
narrations in $\mathcal{N}$. Because $|\mathcal{N}|$ is large, we use a two-stage
procedure: a fast retrieval stage that selects a small candidate pool, followed by
a high-fidelity scorer that re-scores the pool.

Both stages use the same front-end. Each step in $s$ and each of the
$L$ segments in narration $n^{(j)}$ is mapped to a unit-norm embedding
$\mathbf{E}(\cdot)$ by a frozen sentence encoder~\cite{sturua2024jina}.
Their pairwise cosine similarities form a matrix
$\mathbf{W}^{(j)} \in \mathbb{R}^{M \times L}$, with entries
$W^{(j)}_{i,k} = \langle \mathbf{E}(s_i),
\mathbf{E}(n^{(j)}_k) \rangle$.
Narration embeddings are precomputed offline.

\emph{Stage 1 -- monotonic alignment retrieval.}
For each narration $n^{(j)}$, we compute a length-normalized
monotonic-coverage score:
\begin{equation}
    A_{\text{mono}}(s, n^{(j)}) =
    \frac{1}{M}
    \max_{1 \le k_1 \le \cdots \le k_M \le L}
    \sum_{i=1}^{M} W^{(j)}_{i,k_i}.
    \label{eq:monocov}
\end{equation}
This selects a matching narration segment for each step in $s$,
while preserving temporal order. We rank narrations by this score
and retain the top-$K$ as $\mathcal{N}_K(s)$.
The alignment follows the order-aware coverage formulation used in
dynamic time warping (DTW)~\cite{sakoe1978dtw} and step-grounding
methods~\cite{dvornik2021dropdtw,dvornik2023stepformer}.
The dynamic-programming implementation is detailed in
Appendix~\ref{app:alignment}.

\emph{Stage 2 -- high-fidelity scoring.}
For each retained $n^{(j)} \in \mathcal{N}_K(s)$, we compute a global
alignment in the style of Needleman--Wunsch~\cite{needleman1970nw},
using cosine similarities as substitution scores and a fixed gap
penalty $g \le 0$. The alignment follows the recurrence
\begin{equation}
    F_{i,k}
    =
    \max\bigl(
        F_{i-1,k-1} + W^{(j)}_{i,k},
        F_{i-1,k} + g,
        F_{i,k-1} + g
    \bigr).
    \label{eq:nw-alignment}
\end{equation}
This scores the entire narration while penalizing inserted or skipped
steps. Let $A_{\text{NW}}(s,n^{(j)}) \in [0,1]$ denote the
path-length-normalized alignment score (Appendix~\ref{app:alignment}).
The final grounding score is
$A(s,\mathcal{N}) =
\max_{n^{(j)} \in \mathcal{N}_K(s)}
A_{\text{NW}}(s,n^{(j)})$.

\vspace{-1.0em}
\paragraph{History-baseline correction.}
\label{sec:history-baseline}
A naive density-style reward $R(\hat{c}) = A(s_{\text{full}}, \mathcal{N})$ has a known failure mode: $s_{\text{full}}$ inherits whatever alignment the history $h$ already carries, so the policy can inflate the score by paraphrasing $h$ rather than producing a substantive continuation, a familiar form of reward hacking~\cite{skalse2022defining,stiennon2020summarize}. We correct for this by subtracting a history baseline, the alignment that $h$ alone produces against the same retrieved pool, and rewarding the policy only for the improvement attributable to $\hat{c}$. This history-relative correction is designed to discourage continuations
that obtain high alignment scores merely by repeating or paraphrasing
the observed history.

We measure progress \emph{relative} to the remaining gap
$1 - A(s_{\text{hist}}, \mathcal{N})$, which puts sequences with high baseline
alignment on the same scale as those with low baseline alignment. Let
\begin{equation}
    \rho(\hat{c}) \;=\; \frac{A(s_{\text{full}}, \mathcal{N}) - A(s_{\text{hist}}, \mathcal{N})}
    {\max\!\left(1 - A(s_{\text{hist}}, \mathcal{N}),\; \varepsilon\right)},
    \label{eq:rho}
\end{equation}
denote the \emph{relative progress}, with a small
$\varepsilon > 0$ preventing division by zero. 
The base reward is then a
piecewise function of $\rho$ that gates positive reward on a minimum progress
threshold and applies a clipped linear penalty otherwise:
\begin{equation}
    R_{\text{base}}(\hat{c}) \;=\;
    \begin{cases}
        A(s_{\text{full}}, \mathcal{N})
            & \text{if } \rho(\hat{c}) \ge \tau, \\[0.4em]
        \mathrm{clip}\!\big(\alpha\,(\rho(\hat{c}) - \tau),\,-1,\,0\big)
            & \text{otherwise},
    \end{cases}
    \label{eq:reward}
\end{equation}
where $\tau \in (0,1)$ is a progress threshold and $\alpha > 0$ is a penalty slope.
The reward equals the corpus-grounded alignment score whenever the completion meets the progress threshold, and otherwise applies a bounded negative penalty whose magnitude grows linearly with the deficit $\tau - \rho$. This creates an intentional discontinuity at $\rho=\tau$, ensuring that positive reward requires sufficient progress beyond the history.
Figure~\ref{fig:teaser} illustrates how candidate plans are
verified against the instructional corpus and how the resulting
rewards update the policy.
Additional implementation details are reported in Appendix~\ref{app:reward}.

\subsection{Training schemes}
\label{sec:training}

We optimize $\pi_\theta$ with GRPO~\cite{shao2024deepseekmath}: for each prompt we sample $G$ completions $\{\hat{c}^{(g)}\}_{g=1}^G$, score them with $R_{\text{base}}$ (eq.~\ref{eq:reward}), and use the within-group standardized rewards as advantages, with the standard Kullback--Leibler (KL) regularization to a reference policy.

\vspace{-1.0em}
\paragraph{Training schemes.}
We consider three schemes combining a supervised cross-entropy loss $\mathcal{L}_{\text{SFT}}$ on the available continuations with the GRPO loss $\mathcal{L}_{\text{GRPO}}$: \emph{SFT only}, optimizing $\mathcal{L}_{\text{SFT}}$ alone; \emph{RL only}, optimizing $\mathcal{L}_{\text{GRPO}}$ alone from the pretrained checkpoint; and \emph{SFT$\to$RL}, an SFT stage followed by RL from the SFT checkpoint. SFT provides a stable per-token signal that anchors the policy to fluent procedural language but pins it to a single trajectory per example; RL injects a trajectory-level signal from $\mathcal{N}$ that lets the policy place mass on multiple valid orderings. We evaluate these training schemes under different supervision regimes (\S\ref{sec:setup}), including mixtures of annotated and weak supervision (\S\ref{sec:experiments}).

\begin{table}[t]
\centering
\footnotesize

\begin{minipage}[t]{0.27\columnwidth}
\centering
\textbf{Input}\\[4pt]
\hrule\vspace{6pt}

\textbf{Socratic}\\
Frozen VLM converts
video to textual history\\[8pt]

\textbf{Video}\\
Direct video-token
conditioning
\end{minipage}\hfill
\begin{minipage}[t]{0.40\columnwidth}
\centering
\textbf{Supervision}\\[4pt]
\hrule\vspace{6pt}

\textbf{Annotated}\\
Curated procedural datasets\\
$h,c$: human-annotated steps\\[8pt]

\textbf{Weak}\\
Curated procedural datasets\\
$h,c$: VLM-derived steps\\[8pt]

\textbf{Fully unsupervised}\\
HowToCaption corpus\\
$h$: corpus-sampled history\\
No labeled continuation\\[2pt]
\end{minipage}\hfill
\begin{minipage}[t]{0.27\columnwidth}
\centering
\textbf{Training}\\[4pt]
\hrule\vspace{6pt}

\textbf{SFT only}\\
Cross-entropy
optimization\\[8pt]

\textbf{RL only}\\
Corpus-grounded
GRPO\\[8pt]

\textbf{SFT$\to$RL}\\
Sequential SFT
and RL
\end{minipage}

\vspace{6pt}
\caption{\textbf{RECIPE configurations.}
Input, supervision, and training are independent design choices.}
\label{tab:configurations}
\end{table}

\section{Experiments}
\label{sec:experiments}

We organize the empirical study around four questions. We first establish that grounding-based
rewards yield a usable signal across model scales (\S\ref{sec:exp-main}). We then
ask whether \methodname{} outperforms supervised alternatives that rely on either
clean annotations or pseudo-labels extracted from noisy video
(\S\ref{sec:exp-supervision}); whether the grounding reward remains effective when
the policy is trained on weakly-supervised continuations rather than human
annotations (\S\ref{sec:exp-weak}); and whether the same reward generalizes to a
Video input configuration in which the policy directly consumes video tokens
(\S\ref{sec:exp-video}). We further analyze the reward components
(\S\ref{sec:exp-reward}) and procedural diversity
(\S\ref{sec:exp-diversity}). A comparison with prior methods on the
fixed-taxonomy VPA benchmark is reported in Appendix~\ref{app:vpa-results}.

\subsection{Setup}
\label{sec:exp-setup}

\paragraph{Datasets.} Our standard training setup and in-domain evaluation use four annotated procedural datasets, CrossTask~\cite{zhukov2019crosstask}, COIN~\cite{tang2019coin}, CaptainCook4D~\cite{CaptainCook4D}, and EgoProceL~\cite{bansal2022my}. Fully unsupervised training instead draws its training examples from the noisy instructional corpus, without curated procedural annotations. Zero-shot generalization is evaluated on NIV~\cite{alayrac2016unsupervised}, Ego-Exo4D~\cite{grauman2024egoexo4d}, and EgoPER~\cite{leeprocedural2024}, never seen during training. The verification corpus $\mathcal{N}$ is the HowToCaption release~\cite{shvetsova2024howtocaption} of HowTo100M, used as published; details in Appendix~\ref{app:corpus}.
\vspace{-1.0em}
\paragraph{Models and configurations.}
Unless otherwise stated, $\pi_\theta$ is instantiated as a Qwen2.5 text LLM in
the \emph{Socratic} input configuration (\S\ref{sec:inputs}), where a frozen
Qwen2.5-VL-3B serves as the vision-language model $\Phi$. We report results at
three scales: 0.5B, 3B, and 7B parameters. The Base checkpoint is the off-the-shelf Qwen2.5-Instruct model, without fine-tuning. The Video configuration
uses Qwen2.5-VL-3B as the policy and is reported separately in
\S\ref{sec:exp-video}.
\vspace{-1.0em}
\paragraph{Evaluation protocol.}
Generated plans are open-ended natural-language step sequences, which precludes exact-match metrics. We adopt a reference-based LLM-as-judge protocol with Llama-3.3-70B-Instruct. The judge scores each continuation on six $0$--$5$ axes that target distinct failure modes: logical progression (cause-and-effect), temporal sequencing, spatial grounding, continuation (smooth handoff from the history without repetition), clarity, and semantic alignment with the reference. Two penalty rules cap scores when the prediction is verbose or empty (full rubric, prompt, and penalty rules in Appendix~\ref{app:rubric}). We report \emph{macro rubric score} (\%): per-example scores are normalized by the $30$-point maximum, averaged within each dataset, and macro-averaged across the datasets in each split (in-domain or zero-shot). Each prompt is evaluated on a single sampled continuation (Score@$1$). We validate the judge against human preferences on 178 examples (103 in-domain
and 75 zero-shot), with 13 raters comparing continuations from five training
schemes. The judge achieves a system-level Spearman correlation of
$\rho=0.900$ with human rankings, 70.9\% pairwise agreement, and 82.4\%
pairwise agreement after tie calibration. The complete protocol and
per-split results are reported in Appendix~\ref{app:human-eval}.

\subsection{Main results}
\label{sec:exp-main}

\begin{figure}
    \includegraphics[width=1\linewidth]{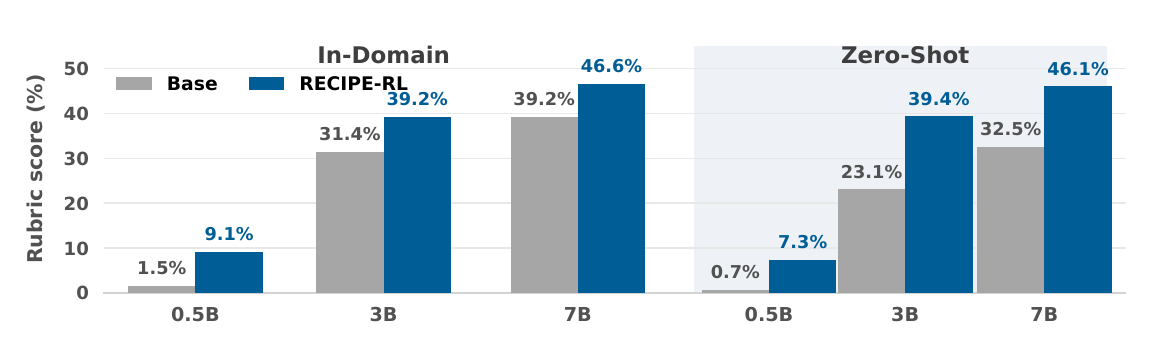}\vspace{-.2cm}
\caption{\textbf{Main results: macro rubric score (\%).} \methodname{}-RL improves over the base checkpoint at every scale and on both splits. All Qwen models use the Socratic configuration and annotated supervision. Per-dataset breakdown in Table~\ref{tab:appendix-perdataset}; closed-source frontier reference in Table~\ref{tab:appendix-frontier}.}
    
    \label{fig:main}
\end{figure}

Figure~\ref{fig:main} summarizes the headline result: \methodname{}-RL improves over the base checkpoint at every scale and on every benchmark, with gains of $+7$ to $+8$ points in-domain and up to $+16$ points zero-shot. At 7B, in-domain and zero-shot macro rubric score are within $0.5$ points ($46.6$ vs.\ $46.1$), indicating the gains do not come at the cost of generalization. The per-criterion breakdown in Appendix~\ref{app:per-criterion} shows that the improvements extend to Logical Progression and Temporal Alignment, rather than being limited to Clarity.

\begin{figure}
    \centering
    \includegraphics[width=1\linewidth]{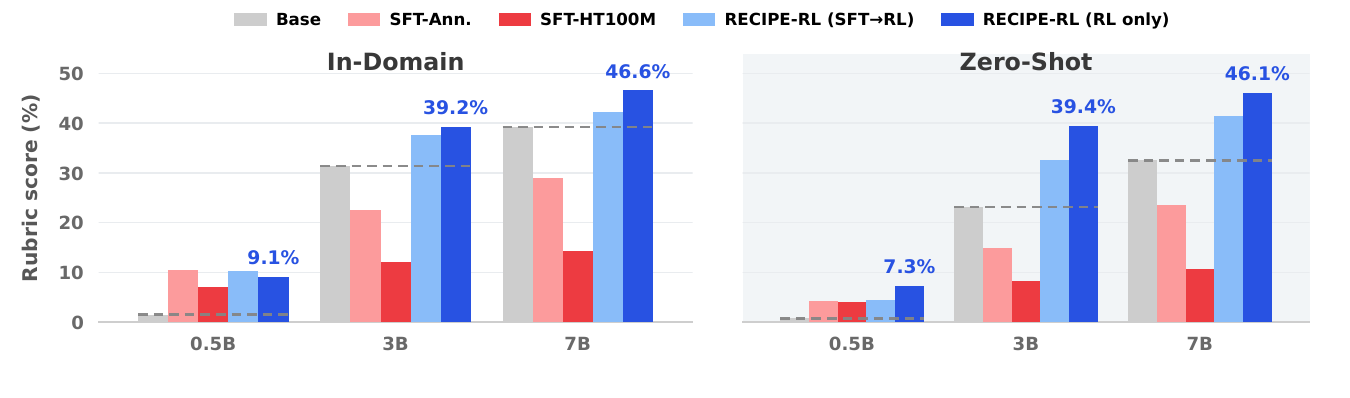}\vspace{-.2cm}
\caption{\textbf{Same corpus, two roles.}
In this comparison, using HT100M pseudo-labels for SFT degrades
performance relative to Base, whereas using the corpus for RL
verification yields substantial improvements. Both \methodname{}
variants (RL only and SFT$\to$RL) outperform both SFT baselines.}
\vspace{-0.5cm}
    
    \label{fig:supervision_results}

\end{figure}

\subsection{\methodname{} outperforms supervised alternatives}
\label{sec:exp-supervision}

We isolate the contribution of the grounding-based RL signal by comparing
\methodname{} against two SFT baselines that use the same training data without
the grounding reward: SFT on the annotated continuations
(\textsc{SFT-Ann.})~and SFT on continuations pseudo-labeled from
HowTo100M~(\textsc{SFT-HT100M}). The latter mirrors the standard pseudo-labeling
recipe in the literature and serves as our most direct test of the
\emph{verification vs.\ labeling} dichotomy: both methods see HowTo100M, but only
\methodname{} uses it as a verifier rather than a labeling source.

Figure~\ref{fig:supervision_results} shows that both \methodname{} variants, RL only
and SFT$\to$RL, outperform the SFT baselines at every scale.
The contrast with \textsc{SFT-HT100M} is stark: pseudo-labeling
\emph{degrades} the base checkpoint at every scale on both splits (e.g.,
$31.4\to12.1$ in-domain at 3B), consistent with the noise-extraction failure
mode that motivates our work. In contrast, RL-only with the same noisy corpus
recovers and exceeds the base by $7.8$ points in-domain and $16.3$ points
zero-shot at 3B. Comparing the two \methodname{} schemes on Qwen, RL-only matches or
edges out SFT$\to$RL at all scales and on both splits. We adopt
\emph{RL-only} as the default \methodname{} training scheme
in subsequent experiments.\vspace{-.1cm}

\paragraph{Generalization across model families.}
We extend the annotated-supervision Socratic experiments to Llama-3.2-3B\cite{grattafiori2024llama} and Gemma-3n-E2B~\cite{team2025gemma}, keeping the datasets, training,
reward, and evaluation setup unchanged. On both backbones,
\methodname{}-RL (RL only) outperforms both SFT baselines in-domain
and zero-shot. Relative to Base, RL-only improves zero-shot
performance from 31.4 to 35.4 on Llama and from 45.8 to 47.4
on Gemma. In-domain, however, its gains over Base are small (+0.2 on Llama, +1.3 on Gemma). The benefit of an SFT stage before RL varies across
backbones: SFT$\to$RL improves Llama in-domain performance
from 33.9 to 38.9 relative to RL-only, but performs worse
zero-shot and on both Gemma splits, where it also falls below Base. Complete results are
reported in Appendix~\ref{app:cross-family}.

\subsection{Robustness to weak supervision}
\label{sec:exp-weak}

A central claim of this work is that the grounding reward reduces the dependence
on clean annotations. We test this by varying the fraction of \emph{weakly}
supervised training examples, in which both the prompt history $h$ and the
continuation $c$ are derived from $\Phi(V)$ rather than from human annotations
(see \S\ref{sec:setup}), from $0\%$ to $100\%$ in $25\%$ increments. The
mixing happens on a \emph{per-example basis}: for each training example we
independently decide whether $(h, c)$ comes from the human-annotated trace or
from the VLM-extracted pseudo-trace, with the assignment held fixed across
epochs.

\begin{figure}[t]
\centering
\begin{minipage}[t]{0.60\linewidth}
    \centering
    \vspace{0pt}
    \includegraphics[width=\linewidth]{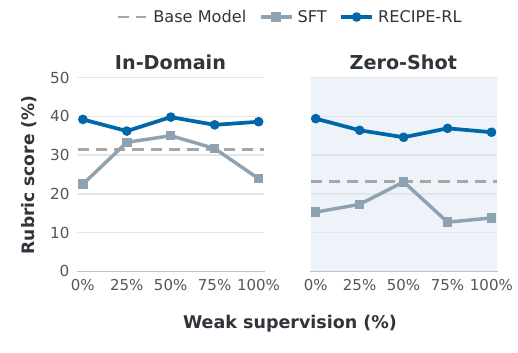}
\caption{\textbf{Robustness to weak supervision (Qwen2.5-3B).}
\methodname{}-RL remains stable across supervision mixes, while SFT degrades
when annotations are scarce, falling below Base in zero-shot evaluation
at $\geq 75\%$ weak supervision.}
\label{fig:p1-robustness}
\end{minipage}%
\hfill
\begin{minipage}[t]{0.38\linewidth}
    \centering
    \vspace{0pt}
    \resizebox{\linewidth}{!}{%
    \setlength{\tabcolsep}{2pt}
    \begin{tabular}{@{}l cc cc cc@{}}
    \toprule
     & \multicolumn{2}{c}{\textbf{Base}} & \multicolumn{2}{c}{\textbf{RL, 0\%}} & \multicolumn{2}{c}{\textbf{RL, 100\%}} \\
    \cmidrule(lr){2-3}\cmidrule(lr){4-5}\cmidrule(lr){6-7}
    \textbf{Model} & In & ZS & In & ZS & In & ZS \\
    \midrule
    Qwen-0.5B &  1.5 &  0.7 &  9.1 &  7.3 & 11.9 &  9.1 \\
    Qwen-3B   & 31.4 & 23.1 & 39.2 & 39.4 & 38.6 & 36.0 \\
    Qwen-7B   & 39.2 & 32.5 & 46.6 & 46.1 & 46.3 & 46.3 \\
    \bottomrule
    \end{tabular}%
    }
\captionof{table}{\textbf{Weak supervision matches annotated training.}
Across model scales, \methodname{}-RL with 100\% weak supervision
performs comparably to training with annotations.
Macro rubric score (\%); ``RL, $p\%$'' denotes $p\%$ weak supervision.}    \label{tab:p1-scale}
\end{minipage}
\end{figure}

Figure~\ref{fig:p1-robustness} reports macro rubric score across the sweep for
Qwen2.5-3B. First, \methodname{}-RL is
\emph{flat} across the supervision mix: in-domain scores stay in the
$36$--$40$ range and zero-shot rubric score in the $35$--$39$ range, regardless of
how much (or how little) clean annotation is available. 
Second, SFT exhibits the
opposite behavior, peaking at intermediate mixes and collapsing at both
extremes, with the zero-shot rubric score falling below the
no-fine-tuning baseline at high weak-supervision fractions
($12.8$ at $25\%$ annotated, $13.7$ at $0\%$ annotated,
vs.\ a base of $23.1$).
SFT faces a tradeoff between the two extremes: clean-only training never sees the noisier $\Phi(h_V)$ histories it must read at test time, while pseudo-label-only training fits noisy text and inherits the noise; \methodname{}-RL avoids it because a noisy pseudo-trace is just a prompt to generate from, and the reward catches mistakes SFT would lock in.

Table~\ref{tab:p1-scale} extends this to $0.5$B and $7$B at the two extremes ($0\%$ and $100\%$ weak). At $0.5$B and $7$B, \methodname{}-RL with fully weak supervision matches or \emph{exceeds} fully annotated training on both splits; at $3$B the $100\%$-weak run is within $0.6$/$3.4$ points (in-domain/zero-shot) of the annotated run. 

An ablation of the maximum training history length is reported
in Appendix~\ref{app:history-length}.

\paragraph{Fully unsupervised training.}
We further remove curated procedural datasets from training.
Histories are sampled as prefixes of HowToCaption narration
sequences. For SFT-HT100M, the corresponding suffix serves as
the supervised continuation target. In contrast,
\methodname{}-RL generates continuations with the policy and
scores them against the HowToCaption corpus, without using
narration suffixes as supervision.
Table~\ref{tab:unsupervised} reports results on the procedural
benchmarks, which are all unseen during training.
\methodname{}-RL improves over Base at every scale, while
SFT on the same corpus improves only the 0.5B model.

\begin{figure}[t]
\centering
\begin{minipage}[t]{0.48\linewidth}
    \centering
    \vspace{0pt}
    \footnotesize
    \setlength{\tabcolsep}{3pt}
    \begin{tabular}{@{}lccc@{}}
    \toprule
    \textbf{Model} & \textbf{Base} & \textbf{SFT-HT100M} &
    \textbf{\methodname{}-RL} \\
    \midrule
    Qwen-0.5B & 1.1 & 6.0 & \textbf{22.6} \\
    Qwen-3B & 29.9 & 19.2 & \textbf{38.1} \\
    Qwen-7B & 36.4 & 22.7 & \textbf{45.1} \\
    \bottomrule
    \end{tabular}
    \vspace{6pt}
\captionof{table}{\textbf{Fully unsupervised training.}
Using only HowToCaption for training, \methodname{}-RL improves over
Base and SFT-HT100M at all three scales on held-out procedural
benchmarks. Macro rubric score (\%).}
    \label{tab:unsupervised}
\end{minipage}%
\hfill
\begin{minipage}[t]{0.48\linewidth}
    \centering
    \vspace{0pt}
    \footnotesize
    \setlength{\tabcolsep}{4pt}
    \begin{tabular}{@{}lcc@{}}
    \toprule
    \textbf{Split} &
    \shortstack{\textbf{Pairwise}\\\textbf{agreement}} &
    \shortstack{\textbf{Calibrated}\\\textbf{accuracy}} \\
    \midrule
    All & 70.9\% & 82.4\% \\
    In-domain & 68.6\% & 80.6\% \\
    Zero-shot & 74.2\% & 85.0\% \\
    \bottomrule
    \end{tabular}
    \vspace{6pt}

\captionof{table}{\textbf{Human evaluation of the LLM judge.}
The judge agrees with human preferences in 70.9\% of pairwise
comparisons, increasing to 82.4\% with tie calibration.} \label{tab:human-eval}
\end{minipage}
\end{figure}

\subsection{Generality across input configurations}
\label{sec:exp-video}

The Video input configuration (\S\ref{sec:inputs}) lets the policy consume video tokens directly. Because Video training is substantially more expensive than the Socratic pipeline, we run a single-backbone study (Qwen2.5-VL-3B) trained on CrossTask only, evaluated on CrossTask (in-domain) and NIV (zero-shot). The numbers in Table~\ref{tab:p2-video} should therefore be read as a targeted test that the grounding reward transfers to direct video-token conditioning, not as a full comparison to the Socratic setting.

Table~\ref{tab:p2-video} confirms that the grounding reward transfers to the
Video configuration: \methodname{}-RL improves over Base on NIV with both
annotated and weakly supervised continuations ($+0.6$ and $+1.1$ points
respectively), while SFT \emph{degrades} the base checkpoint on NIV by $7.9$
points, the same SFT-collapse pattern observed in the Socratic
configuration. CrossTask gains are smaller and noisier in this single-scale
study and we are cautious about reading them too closely.

We also evaluate \methodname{} on the more constrained setting of
fixed-taxonomy planning (VPA), with results and comparisons to
existing methods reported in Appendix~\ref{app:vpa-results}.

\begin{table}[t]
\centering
\begin{minipage}[t]{0.43\linewidth}
    \centering
    \vspace{0pt}
    \resizebox{\linewidth}{!}{%
    \setlength{\tabcolsep}{4pt}
    \begin{tabular}{@{}l cc@{}}
    \toprule
    \shortstack{\textbf{Training scheme}\\~} & \makecell{\textbf{CrossTask}\\\textbf{(in-domain)}} & \makecell{\textbf{NIV}\\\textbf{(zero-shot)}} \\
    \midrule
    Base                                    & 25.5 & 36.5 \\
    SFT (annotated)                         & 28.0 & 28.6 \\
    SFT (weak)                              & 27.9 & 28.1 \\
    \methodname{}-RL (annotated)            & 27.2 & 37.1 \\
    \methodname{}-RL (weak)                 & 26.1 & 37.6 \\
    \bottomrule
    \end{tabular}%
    }
    \vspace{6pt}
    \captionof{table}{\textbf{Reward transfers to the Video configuration.} With Qwen2.5-VL-3B reading video tokens directly, \methodname{}-RL improves over the base under both annotated and weak supervision; SFT degrades the zero-shot benchmark, mirroring the Socratic SFT-collapse pattern.}
    \label{tab:p2-video}
\end{minipage}%
\hfill
\begin{minipage}[t]{0.55\linewidth}
    \centering
    \vspace{0pt}
    \resizebox{\linewidth}{!}{%
    \setlength{\tabcolsep}{4pt}
    \begin{tabular}{@{}cccc cc@{}}
    \toprule
    \multicolumn{4}{c}{\textbf{Reward Components}} & \multicolumn{2}{c}{\textbf{macro rubric score (\%)}} \\
    \cmidrule(lr){1-4} \cmidrule(lr){5-6}
    \shortstack{\textbf{Stage 2}\\~} & \makecell{\textbf{History}\\\textbf{baseline}} & \makecell{\textbf{Relative}\\\textbf{progress}} & \makecell{\textbf{Progress}\\\textbf{gate}} & \shortstack{\textbf{In-domain}\\~} & \shortstack{\textbf{Zero-shot}\\~} \\
    \midrule
    $\times$   & \checkmark & \checkmark & \checkmark & 32.0 & 28.8 \\
    \checkmark & $\times$   & \checkmark & \checkmark & 33.4 & 30.8 \\
    \checkmark & \checkmark & $\times$   & \checkmark & 30.9 & 33.7 \\
    \checkmark & \checkmark & \checkmark & $\times$   & 35.4 & 34.0 \\
    \midrule
    \checkmark & \checkmark & \checkmark & \checkmark & \textbf{39.2} & \textbf{39.4} \\
    \bottomrule
    \end{tabular}%
    }
    \vspace{6pt}
    \captionof{table}{\textbf{Every reward component pulls weight.} Removing any one of the four reward components costs $4$--$11$ macro rubric score points; Stage 2 high-fidelity scoring is the most critical for zero-shot generalization. Qwen2.5-3B (\%).}
    \label{tab:p5-ablation}
\end{minipage}
\end{table}

\subsection{Reward-component analysis}
\label{sec:exp-reward}

We isolate each reward component on Qwen2.5-3B (Table~\ref{tab:p5-ablation}). Each removal hurts on both splits in a way that maps to the failure mode the component was designed to prevent: dropping Stage 2 high-fidelity scoring is most damaging ($-7$/$-11$, since Stage 1 alone is too coarse to discriminate); removing the history baseline costs $6$/$9$ points (the policy paraphrases the history to inflate reward); dropping relative-progress normalization ($-8$/$-6$) inflates rewards on already-aligned prompts; 
and removing the progress gate ($-4$/$-5$) lets marginally-improving completions earn positive rewards. Appendix~\ref{app:reward-fidelity} further compares the grounding reward with the LLM judge, including cases where their assessments disagree.

\Needspace{24\baselineskip}

\subsection{Procedural diversity}
\label{sec:exp-diversity}

\begin{wrapfigure}{r}{0.48\linewidth}
\vspace{-4pt}
\centering
\includegraphics[width=\linewidth]{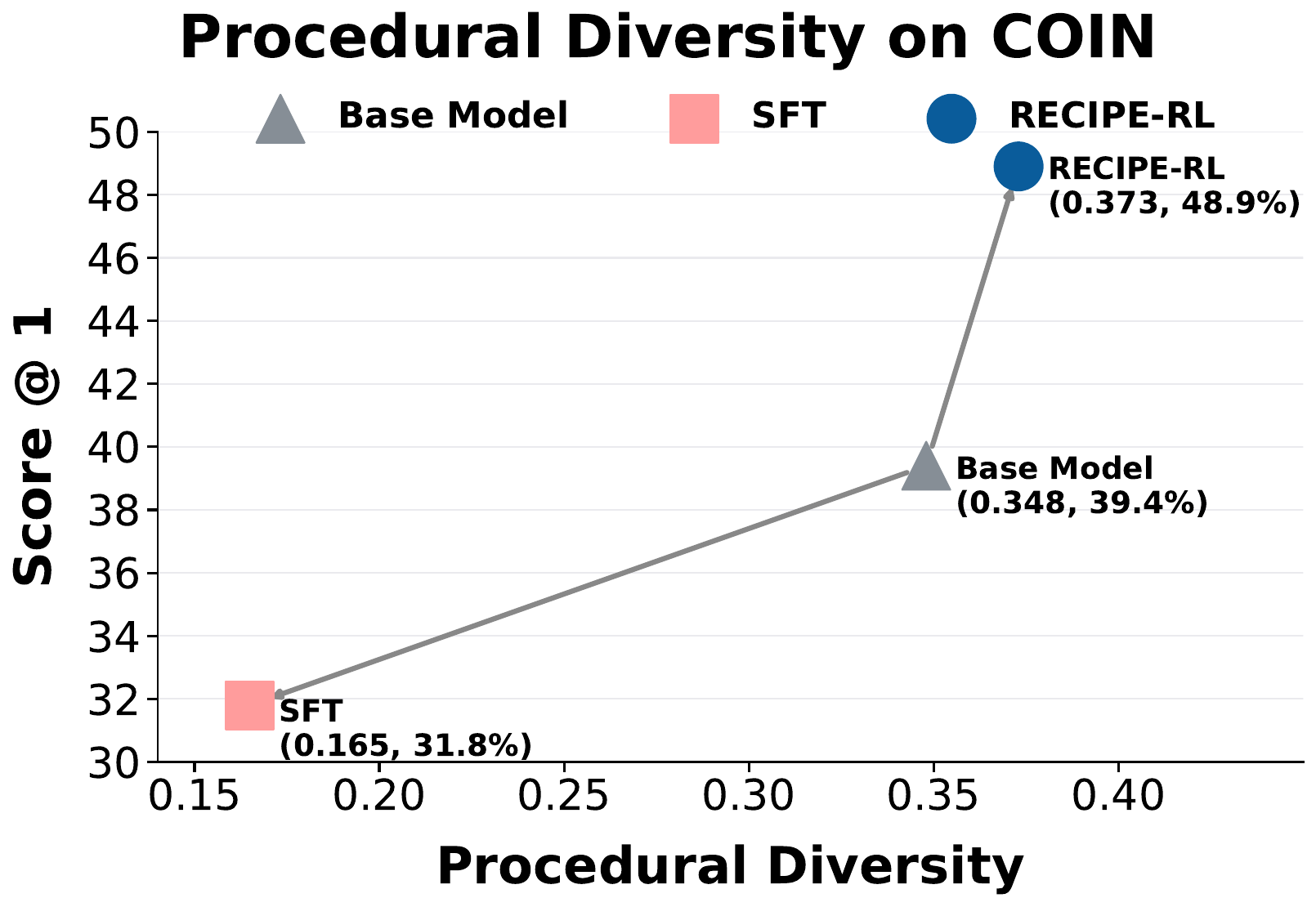}
\vspace{0pt}
\caption{\textbf{\methodname{}-RL is both more diverse and more accurate than SFT.}
Procedural diversity (x-axis) vs.\ Score@$1$ macro rubric score (y-axis)
on COIN; \methodname{}-RL is the only configuration in the upper-right region.}
\label{fig:procedural-diversity}
\vspace{-8pt}
\end{wrapfigure}

A reward that accommodates \emph{many} valid orderings (\S\ref{sec:techapproach}) should leave the policy free to express that variety, whereas a single-reference SFT objective should push the policy toward one canonical plan. We test this hypothesis on $100$ prompts from the COIN test split: for each prompt we draw $10$ generations from each model and use Gemini-3.1-Pro as a pairwise judge to assign a procedural-diversity score to the resulting set (judge prompt and aggregation in Appendix~\ref{app:diversity}). 
Figure~\ref{fig:procedural-diversity} plots procedural diversity against the Score@$1$ macro rubric score.

The three configurations land in three distinct regions of the plot. SFT collapses on \emph{both} axes: it produces the least diverse generations (procedural diversity $0.165$, less than half of the Base checkpoint) \emph{and} the lowest rubric score ($31.8$). The Base checkpoint preserves diversity ($0.348$) but at a modest rubric score ($39.4$). \methodname{}-RL is the only configuration that lands in the upper-right of the plot, matching the Base checkpoint's diversity ($0.373$) while adding nearly $10$ rubric-score points ($48.9$). The grounding reward therefore preserves the natural variability of valid plans rather than collapsing it onto a single annotated trajectory, supporting the design intuition behind using the corpus as a verifier rather than a labeling source.

\section{Conclusion}

\methodname{} reframes a noisy video corpus as a verifier of generated
plans rather than a source of training labels. Across three model
scales, it improves over the base checkpoints and outperforms SFT
on annotated and pseudo-labeled continuations. These improvements
extend to fully unsupervised training, while human evaluation
supports the reliability of our LLM-as-judge protocol. Unlike SFT,
\methodname{}-RL also preserves procedural diversity.
Single-reference evaluation and text-only verification remain
limitations (Appendix~\ref{app:limitations}).

\section*{Acknowledgements and Disclosures}

This work was supported in part by Northeastern University startup
funds and the President Joseph E. Aoun Chair. We acknowledge the
computational resources and support provided by Northeastern
University's Research Computing team through the Explorer Cluster
and by CINECA through the ISCRA initiative. We thank Shih-Po Lee
for useful discussions. The authors declare no competing interests.

{
\small
\bibliographystyle{plain}
\bibliography{bibtex}

}

\appendix

\section{Implementation details}
\label{app:implementation}

\subsection{HowToCaption corpus}
\label{app:corpus}

The reward signal is computed against the HowToCaption release of HowTo100M~\cite{shvetsova2024howtocaption}, which reprocesses raw ASR subtitles into caption-like text via Vicuna-13B and applies automatic alignment and filtering, yielding $\sim$25M video-text pairs from 1.2M instructional videos. We use this release as published, with no additional filtering, deduplication, or resegmentation: captions are segmented by their original timestamps, and each segment is converted to a sentence embedding with the frozen Jina encoder~\cite{sturua2024jina}. The corpus index, the precomputed embeddings, and the indexing scripts are released with the code.

\subsection{Two-stage corpus alignment}
\label{app:alignment}

\paragraph{Embeddings.}
Each step in $s = (s_1, \ldots, s_M)$ and each narration segment in $n^{(j)} = (n^{(j)}_1, \ldots, n^{(j)}_L)$ is independently mapped to a unit-norm $d$-dimensional vector by the frozen sentence encoder of~\cite{sturua2024jina}, following the sentence-BERT framework~\cite{reimers2019sbert}. This gives step-level embedding sequences $\mathbf{E}(s) \in \mathbb{R}^{M \times d}$ and $\mathbf{E}(n^{(j)}) \in \mathbb{R}^{L \times d}$, and a cosine-similarity matrix $\mathbf{W}^{(j)} \in \mathbb{R}^{M \times L}$ with entries $W^{(j)}_{i,k} = \langle \mathbf{E}(s)_i,\, \mathbf{E}(n^{(j)})_k \rangle \in [-1, 1]$. All narration embeddings are precomputed once over $\mathcal{N}$.

\paragraph{Stage 1: monotonic-coverage retrieval.}
The score $A_{\text{mono}}$ defined in Eq.~\ref{eq:monocov}
is computed in $O(ML)$ time per pair via the forward
dynamic-programming recursion
\begin{equation}
    D_{i,k} \;=\; W^{(j)}_{i,k} + \max_{k' \le k} D_{i-1,\,k'}, \qquad D_{0,k} := 0 \text{ for all } k,
\end{equation}
giving $A_{\text{mono}}(s, n^{(j)}) = \max_k D_{M,k} / M$. We sort the corpus by $A_{\text{mono}}(s, \cdot)$ and retain the top-$K$ as $\mathcal{N}_K(s)$, with $K = 25$.

\paragraph{Stage 2: Needleman--Wunsch on cosine similarities.}
The $O(ML)$ Needleman--Wunsch recurrence is given in
Eq.~\ref{eq:nw-alignment}. After computing the optimal global-alignment score $F_{M,L}$, we backtrace one optimal alignment path $\pi^\star$ from $(M,L)$ to $(0,0)$. Let $|\pi^\star|$ denote the number of alignment steps in this path, where diagonal, vertical, and horizontal moves each count as one step. We define the normalized Needleman--Wunsch score as
\begin{equation}
    A_{\text{NW}}(s,n^{(j)})
    =
    \operatorname{clip}\!\left(
        \frac{F_{M,L}}{\max(|\pi^\star|,1)},
        10^{-6},
        1
    \right).
\end{equation}
We use $g=-0.05$.

\subsection{Reward function}
\label{app:reward}

\paragraph{Hyperparameters.}
We use $\tau = 0.10$ for the progress threshold, $\alpha = 2.0$ for the penalty slope, and $\varepsilon = 10^{-6}$ as the zero-division guard in eq.~\ref{eq:rho}.

\paragraph{Reward optimization.}
We did not observe instability arising from the discontinuity at $\rho=\tau$ in any of our runs. In practice, the policy moves above the threshold within the first few thousand steps.

\subsection{Training procedure}
\label{app:training}

\paragraph{GRPO.}
For each prompt we sample $G = 4$ completions, compute $R_{\text{base}}$ for each, and form advantages $\hat{A}^{(g)} = \big(R_{\text{base}}^{(g)} - \mu\big) / \sigma$ where $\mu, \sigma$ are the within-group mean and standard deviation. The GRPO objective adds a KL penalty $\beta \cdot \mathrm{KL}\big(\pi_\theta \| \pi_{\text{ref}}\big)$ with $\beta = 0.04$ and $\pi_{\text{ref}}$ the SFT or pretrained checkpoint depending on the regime. We use clipping ratio $\epsilon_{\text{clip}} = 0.30$.

\paragraph{SFT.}
Standard token-level cross-entropy on $(h, c)$ pairs as defined in \S\ref{sec:setup}, with batch size $2$ and learning rate $2 \times 10^{-4}$ for $3$ epochs.

\paragraph{Optimizer and schedule.}
For both supervised fine-tuning (SFT) and reinforcement learning (RL), we use the AdamW optimizer \cite{loshchilov2018decoupled}, leveraging its fused implementation for computational efficiency. During SFT, we employ a cosine learning rate schedule with a peak learning rate of $2 \times 10^{-4}$ and a warmup ratio of 10\%. Conversely, for the RL stage, we maintain a constant learning rate of $1 \times 10^{-4}$ with no warmup steps.

\subsection{Compute Resources}
\label{app:compute_resources}
The experiments were run on a server equipped with eight NVIDIA H200 GPUs, of which we used four for this work. Two GPUs were allocated to training, while the remaining two were used for evaluation, allowing model optimization, rollout generation, reward computation, and LLM-as-judge evaluation to be partially parallelized. The full experimental pipeline required approximately two days. The most computationally demanding stage was \methodname{}-RL training, because each optimization step requires sampling multiple candidate continuations and computing grounding-based rewards for them. Supervised fine-tuning was comparatively less expensive, while the evaluation stage introduced additional inference cost but could be distributed across the evaluation GPUs.

\section{Evaluation protocol}
\label{app:rubric}

We score each generated continuation against the held-out reference using Llama-3.3-70B-Instruct as the judge. The judge receives the goal, the textual history, the reference (gold) continuation, and the predicted continuation, and returns integer scores from $0$ to $5$ on six axes. The maximum total is $30$ points per example, which we report normalized to a percentage as macro rubric score in the main text.

\paragraph{Penalty rules.} Two rules are applied before per-criterion scoring. A \emph{fluff penalty} caps Semantic Alignment, Clarity, and the Holistic score at $2$ when the prediction adds unnecessary words, generic advice, or descriptive adjectives not present in the reference; numbering and bullet formatting are exempt. A \emph{fatal-failure} rule sets all scores to $0$ when the prediction is empty or contains no actionable steps.

\paragraph{Criteria.} Each criterion uses the same $0$--$5$ scale, where $5$ means perfect alignment with the reference and $0$ means severe contradiction.
\begin{itemize}[leftmargin=*, topsep=0.2em, itemsep=0.2em]
\item \textbf{Logical progression.} Strict cause-and-effect: each step has its physical prerequisite from earlier steps. Penalizes unstated assumptions and contradictions of the reference's logical chain.
\item \textbf{Temporal alignment.} Correct chronology and sequencing. Penalizes unprompted insertions, reorderings, and verbosity that introduces temporal drift.
\item \textbf{Spatial grounding.} Physical and spatial coherence with the visible scene. Penalizes hallucinated objects or locations that contradict the reference.
\item \textbf{Continuation.} Smooth handoff from the history. Penalizes repetition of history steps, abrupt transitions, and skipping necessary bridging steps.
\item \textbf{Clarity.} Precision and readability. Penalizes verbose phrasing that buries the core instruction.
\item \textbf{Semantic alignment.} Match with the reference's intent and overall plan. Penalizes off-path predictions that introduce different objects or actions.
\end{itemize}

\paragraph{Output and parsing.} The judge is instructed to return only a JSON block with one integer per criterion. To handle occasional malformed output, we use a fallback regex parser over key-value pairs, with extracted values clamped to $[0,5]$. The full judge prompt is included in the released code.

\subsection{Procedural-diversity metric}
\label{app:diversity}

To quantify whether a model produces genuinely different valid continuations,
rather than superficial paraphrases, we use an LLM-based pairwise procedural-diversity
metric. For each prompt, we sample $N=10$ continuations from the model. We then
compare all unordered pairs of generations for the same prompt, yielding
$\binom{10}{2}=45$ pairwise comparisons per prompt.

Each pair is evaluated by Gemini-3.1-Pro with a goal-aware diversity prompt. The
judge receives the task goal, the history, and two predicted plans.
It is asked to decide whether the two plans describe the same procedure or a
different procedure, while explicitly considering whether both plans are plausible
and useful for achieving the goal. This distinction is important: generations are
not rewarded for being different if the difference comes from degeneration,
repetition, irrelevance, or failure to progress toward the goal. The judge returns
a discrete diversity score in $\{0.0, 0.25, 0.5, 0.75, 1.0\}$, where $0.0$
indicates paraphrases or near-identical procedures, $0.25$ indicates surface-level
differences caused by weak or degenerate outputs, $0.5$ indicates partial
goal-related differences, and $0.75$ or $1.0$ indicates clearly different
goal-affine procedural stages or action sequences.

For a prompt $x$, let $\hat{c}_1,\ldots,\hat{c}_{10}$ denote the sampled
continuations and let $d(\hat{c}_i,\hat{c}_j)$ be the Gemini diversity score for
the pair $(i,j)$. The procedural-diversity score for the prompt is the mean
pairwise score:
\[
D(x) =
\frac{2}{N(N-1)}
\sum_{1 \leq i < j \leq N}
d(\hat{c}_i,\hat{c}_j),
\qquad N=10.
\]
The model-level diversity reported on the $x$-axis of
Figure~\ref{fig:procedural-diversity} is obtained by averaging $D(x)$ over
100 randomly selected COIN prompts.

The prompt used for Gemini-3.1-Pro enforces three main rules. First, semantic
procedural intent is compared rather than exact wording, so synonyms,
formatting, numbering, and harmless stylistic differences are ignored. Second,
repeated redundant steps are collapsed before comparison, preventing repetition
from being counted as useful diversity. Third, high diversity requires both
procedural difference and goal affinity: if one plan is off-topic, nonsensical,
repetitive, or not useful for completing the goal, the diversity score is capped
at a low value. This makes the metric closer to ``useful procedural diversity''
than to lexical or embedding-level dispersion.

For Figure~\ref{fig:procedural-diversity}, we evaluate 100 COIN prompts and
generate 10 continuations per prompt for each model. The corresponding quality
score on the $y$-axis is Score@1, computed using the same reference-based
evaluation protocol used in the main experiments. This analysis shows that SFT
collapses procedural variability, producing low diversity and low quality,
whereas \methodname{}-RL improves both diversity and Score@1 relative to the
base model.

\section{Additional results}
\label{app:results}

\subsection{Human evaluation of the LLM judge}
\label{app:human-eval}

We evaluate the agreement between the LLM judge and human preferences
on 178 examples (103 in-domain and 75 zero-shot). Thirteen human
raters compare the continuations generated by Base, SFT-Ann.,
SFT-HT100M, \methodname{} (SFT$\to$RL), and
\methodname{}-RL, given the goal, history, and reference continuation.
For each example, raters select their preferred continuation.

We measure system-level Spearman correlation over the rankings of
the five methods and pairwise agreement between the judge and human
preferences. The Spearman correlation is $\rho=0.900$ on both splits.
Pairwise agreement is 70.9\% overall, increasing to 82.4\% after
tie calibration (Table~\ref{tab:human-eval}).

The judge assigns scores on six discrete $0$--$5$ criteria, with a
maximum of 30 points. In 56.7\% of queries, the difference between
the judge scores of the human-preferred continuation and the nearest
alternative is at most one point. We therefore also report
$\epsilon$-calibrated pairwise accuracy, treating score differences of at most $\epsilon=1/30$ (one raw point) as ties.

\subsection{Comparison on Visual Planning for Assistance}
\label{app:vpa-results}

We evaluate \methodname{} on the Visual Planning for Assistance (VPA)
benchmark~\cite{patel2023pretrained,islam2024propose} using the standard
CrossTask splits at $T \in \{1,3,4\}$. Here $T$ is the planning horizon (number of steps to predict). We report mean accuracy (mAcc), success rate (SR), and mean intersection over union (mIoU). Since our policy generates
free-form natural-language steps, predictions are mapped to the
closest admissible action in the CrossTask taxonomy of 105 actions
using embedding-based nearest-neighbor matching. This evaluation
therefore measures performance in a fixed-taxonomy setting rather
than the open-vocabulary setting of our main experiments.

\paragraph{Comparison with prior methods.}
Table~\ref{tab:vpa-comparison} compares fully supervised methods
and methods without supervised CrossTask continuation targets.
VideoPlan~\cite{Zhang_2026_WACV} generates free-form plans and maps
them to action labels, but uses fully supervised training.
Our annotated-supervision configurations use CrossTask videos to
construct training examples and retrieve corpus narrations, but
do not use their annotated continuations as supervised targets.
They are therefore not strictly zero-shot with respect to
CrossTask training videos.

We integrate \methodname{}-RL into VidAssist's
propose--assess--search pipeline as an additional proposer alongside
the original Llama-2-70B (Appendix~\ref{app:vpa-integration}).
Pairing VidAssist with \methodname{}-RL improves over the original
VidAssist on mAcc and SR at all three horizons. The original
VidAssist achieves higher mIoU at $T=3$ and $T=4$.
VidAssist+Base performs worse than the original VidAssist
on all reported metrics.

\begin{table}[t]
\centering
\small
\setlength{\tabcolsep}{4pt}
\begin{tabular}{@{}l c ccc ccc@{}}
\toprule
\multirow{2}{*}{\textbf{Method}}
& \multicolumn{1}{c}{$T=1$}
& \multicolumn{3}{c}{$T=3$}
& \multicolumn{3}{c}{$T=4$} \\
\cmidrule(lr){2-2}
\cmidrule(lr){3-5}
\cmidrule(lr){6-8}
& \textbf{mAcc}
& \textbf{SR} & \textbf{mAcc} & \textbf{mIoU}
& \textbf{SR} & \textbf{mAcc} & \textbf{mIoU} \\
\midrule
\multicolumn{8}{l}{\textbf{Fully Supervised}} \\
Most Probable & 10.4 & 1.7 & 6.1 & 9.9 & 1.3 & 5.5 & 13.9 \\
Most Probable w/ goal & 12.4 & 2.4 & 8.9 & 15.5 & 1.5 & 7.9 & 20.5 \\
DDN~\cite{chang2020procedure}
 & \underline{33.4} & 6.8 & 25.8 & 35.2 & 3.6 & 24.1 & 37.0 \\
LTA~\cite{grauman2022ego4d}
 & 26.9 & 2.4 & 24.0 & 35.2 & 1.2 & 21.7 & 36.8 \\
VLaMP~\cite{patel2023pretrained}
 & \textbf{50.6} & \underline{10.3} & \underline{35.3}
 & \underline{44.0} & \underline{4.4} & \underline{31.7}
 & \underline{43.4} \\
VideoPlan~\cite{Zhang_2026_WACV}
 & -- & \textbf{15.4} & \textbf{39.4} & \textbf{51.4}
 & \textbf{9.9} & \textbf{37.4} & \textbf{54.3} \\
\midrule
\multicolumn{8}{l}{\textbf{Zero-Shot / No Supervised CrossTask Targets}} \\
Random & 0.9 & 0.0 & 0.9 & 1.5 & 0.0 & 0.9 & 1.9 \\
Random w/ goal & 13.2 & 0.3 & 13.4 & 23.6 & 0.0 & 12.7 & 27.8 \\
LLM Baseline~\cite{touvron2023llama}
 & 25.8 & 2.1 & 23.2 & 27.7 & 0.4 & 18.9 & 33.0 \\
LLM Agent Baseline~\cite{huang2022language}
 & 28.7 & 3.1 & 25.6 & 29.9 & 0.8 & 20.0
 & \underline{35.6} \\
VidAssist~\cite{islam2024propose}
 & \underline{38.7} & \underline{8.7} & \underline{28.5}
 & \textbf{44.1} & \underline{4.6} & \underline{25.8}
 & \textbf{46.8} \\
VidAssist w/ Base
 & 36.5 & 2.0 & 27.3 & 33.0 & 1.0 & 24.3 & 32.0 \\
VidAssist w/ \methodname{}-RL
 & \textbf{39.9} & \textbf{9.3} & \textbf{29.9}
 & \underline{35.7} & \textbf{4.8} & \textbf{26.0}
 & 34.5 \\
\bottomrule
\end{tabular}
\caption{\textbf{Comparison on VPA.}
\methodname{}-RL improves VidAssist on mAcc and SR at every horizon.
Fully supervised VideoPlan achieves higher scores at $T=3$ and $T=4$.
Best in bold and second best underlined within each block.}
\label{tab:vpa-comparison}
\end{table}

\paragraph{Standalone and fully supervised variants.}
Table~\ref{tab:vpa-standalone} reports the additional experiments
without VidAssist's search pipeline and with SFT followed by RL
using 100\% annotated supervision. Standalone \methodname{}-RL
improves over Base on all reported metrics. The fully supervised
SFT$\to$RL variant obtains higher mAcc at $T=4$, but lower mAcc
at $T=1$ than standalone \methodname{}-RL.

\begin{table}[t]
\centering
\small
\setlength{\tabcolsep}{5pt}
\begin{tabular}{@{}l c ccc ccc@{}}
\toprule
\multirow{2}{*}{\textbf{Method}}
& \multicolumn{1}{c}{$T=1$}
& \multicolumn{3}{c}{$T=3$}
& \multicolumn{3}{c}{$T=4$} \\
\cmidrule(lr){2-2}
\cmidrule(lr){3-5}
\cmidrule(lr){6-8}
& \textbf{mAcc}
& \textbf{SR} & \textbf{mAcc} & \textbf{mIoU}
& \textbf{SR} & \textbf{mAcc} & \textbf{mIoU} \\
\midrule
Standalone Base 7B
 & 31.3 & 2.0 & 22.4 & 24.6 & 0.0 & 17.5 & 23.3 \\
Standalone \methodname{}-RL 7B
 & 33.3 & 4.9 & 25.0 & 28.6 & 1.0 & 22.1 & 31.0 \\
SFT$\to$RL (100\% annotated) 7B
 & 23.5 & 3.9 & 22.2 & 25.7 & 1.9 & 23.6 & 31.1 \\
\bottomrule
\end{tabular}
\caption{\textbf{Standalone and fully supervised VPA results.}
Standalone \methodname{}-RL improves over Base at all horizons.
SFT$\to$RL achieves higher mAcc at $T=4$ but lower mAcc at $T=1$.}
\label{tab:vpa-standalone}
\end{table}

\subsection{VidAssist integration}
\label{app:vpa-integration}

VidAssist is an inference-time propose--assess--search procedure: at each step, an LLM proposes candidate next actions and an LLM assessor scores them. We use \methodname{} as an additional proposer and leave the rest of the pipeline untouched. At each step we draw two candidates from VidAssist's original Llama-2-70B proposer and two from \methodname{}-RL (Qwen2.5-7B), project both sets onto the CrossTask taxonomy, and let the same Llama assessor select the best. The selected action is appended to the running plan and the procedure repeats until the planning horizon $T$ is reached.

\subsection{Generalization across model families}
\label{app:cross-family}

Table~\ref{tab:cross-family} reports the results of extending
the annotated-supervision Socratic experiments to Llama-3.2-3B
and Gemma-3n-E2B. The frozen Qwen2.5-VL-3B used to extract
histories is unchanged. Both \methodname{} training schemes
outperform SFT-Ann. in-domain on both backbones, while RL-only
also improves over Base on both zero-shot splits. SFT$\to$RL
performs best in-domain on Llama, whereas RL-only performs
best on both Gemma splits.

\begin{table}[t]
\centering
\small
\setlength{\tabcolsep}{6pt}
\begin{tabular}{@{}llcc@{}}
\toprule
\textbf{Backbone} & \textbf{Training scheme} &
\textbf{In-domain} & \textbf{Zero-shot} \\
\midrule
\multirow{5}{*}{Llama-3.2-3B}
 & Base & 33.7 & 31.4 \\
 & SFT-Ann. & 29.9 & 12.2 \\
 & SFT-HT100M & 11.9 & 7.5 \\
 & \methodname{} (SFT$\to$RL) & \textbf{38.9} & 27.0 \\
 & \methodname{} (RL only) & 33.9 & \textbf{35.4} \\
\midrule
\multirow{5}{*}{Gemma-3n-E2B}
 & Base & 43.5 & 45.8 \\
 & SFT-Ann. & 36.2 & 19.6 \\
 & SFT-HT100M & 20.5 & 10.2 \\
 & \methodname{} (SFT$\to$RL) & 37.9 & 19.6 \\
 & \methodname{} (RL only) & \textbf{44.8} & \textbf{47.4} \\
\bottomrule
\end{tabular}
\caption{\textbf{Generalization across model families.}
\methodname{}-RL outperforms pure SFT on both Llama and Gemma,
although the preferred RL training scheme varies across backbones.
Results are macro-averaged across the in-domain and zero-shot
benchmarks.}
\label{tab:cross-family}
\end{table}

\subsection{Per-criterion evaluation}
\label{app:per-criterion}

Table~\ref{tab:per-criterion} reports the six rubric criteria
separately for the 3B and 7B models. \methodname{}-RL (RL only)
obtains the highest scores on all six criteria at both scales.
The improvements include Logical Progression, Temporal Alignment,
and Continuation, rather than being limited to Clarity.

\begin{table}[t]
\centering
\small
\setlength{\tabcolsep}{4pt}
\begin{tabular}{@{}llcccccc@{}}
\toprule
\textbf{Model} & \textbf{Training scheme} &
\shortstack{\textbf{Logical}\\\textbf{Progr.}} &
\shortstack{\textbf{Temporal}\\\textbf{Align.}} &
\shortstack{\textbf{Spatial}\\\textbf{Ground.}} &
\textbf{Continuation} &
\textbf{Clarity} &
\shortstack{\textbf{Semantic}\\\textbf{Align.}} \\
\midrule
\multirow{5}{*}{3B}
 & Base & 1.26 & 1.10 & 1.72 & 1.47 & 2.24 & 1.19 \\
 & SFT-Ann. & 0.79 & 0.68 & 1.06 & 0.94 & 1.54 & 0.82 \\
 & SFT-HT100M & 0.35 & 0.30 & 0.50 & 0.43 & 1.22 & 0.34 \\
 & \methodname{} (SFT$\to$RL) & \underline{1.52} & \underline{1.36} & \underline{2.04} & \underline{1.88} & \underline{2.46} & \underline{1.41} \\
 & \methodname{} (RL only) & \textbf{1.70} & \textbf{1.50} & \textbf{2.27} & \textbf{2.03} & \textbf{2.68} & \textbf{1.60} \\
\midrule
\multirow{5}{*}{7B}
 & Base & 1.60 & 1.44 & 2.08 & 1.90 & 2.44 & 1.45 \\
 & SFT-Ann. & 1.15 & 0.98 & 1.43 & 1.32 & 1.96 & 1.16 \\
 & SFT-HT100M & 0.46 & 0.40 & 0.62 & 0.51 & 1.41 & 0.45 \\
 & \methodname{} (SFT$\to$RL) & \underline{1.90} & \underline{1.66} & \underline{2.43} & \underline{2.26} & \underline{2.60} & \underline{1.75} \\
 & \methodname{} (RL only) & \textbf{2.07} & \textbf{1.87} & \textbf{2.67} & \textbf{2.47} & \textbf{2.90} & \textbf{1.92} \\
\bottomrule
\end{tabular}
\caption{\textbf{Per-criterion evaluation.}
\methodname{}-RL improves all six procedural-quality criteria
over Base and SFT at both model scales.
Scores are macro-averaged across datasets on a 0--5 scale.
Best in bold, second best underlined.}
\label{tab:per-criterion}
\end{table}

\subsection{Sensitivity to procedural-order perturbations}
\label{app:temporal-stress}

We evaluate the sensitivity of the LLM judge to procedural-order
errors by corrupting ground-truth continuations. Each step is
independently deleted, swapped with its neighbor, or moved to the
end with probability $\alpha$. The corrupted continuations are
scored against the original references. Table~\ref{tab:temporal-stress}
reports the resulting rubric scores and their Pearson correlation
with edit distance from the original continuation. The score
decreases from 75.6\% to 16.4\% as the corruption probability
increases, with negative correlations at all noise levels.

\begin{table}[t]
\centering
\small
\setlength{\tabcolsep}{8pt}
\begin{tabular}{@{}ccc@{}}
\toprule
\textbf{Noise probability} &
\textbf{Macro rubric score} &
\shortstack{\textbf{Pearson correlation}\\
\textbf{with edit distance}} \\
\midrule
$\alpha=0.10$ & 75.6\% & $-0.71$ \\
$\alpha=0.25$ & 55.7\% & $-0.68$ \\
$\alpha=0.50$ & 37.7\% & $-0.60$ \\
$\alpha=0.75$ & 23.8\% & $-0.66$ \\
$\alpha=0.90$ & 16.4\% & $-0.72$ \\
\bottomrule
\end{tabular}
\caption{\textbf{Sensitivity to procedural-order perturbations.}
Judge scores decrease as corruption increases, with negative
correlations between score and edit distance at every noise level.}
\label{tab:temporal-stress}
\end{table}

\subsection{Valid alternative and infeasible plans}
\label{app:valid-alternatives}

We further evaluate whether the judge distinguishes valid alternative
orderings from infeasible plans. We construct two subsets of 50 corrupted
continuations each, manually verifying whether the ordering preserves
the procedural prerequisites. Valid alternatives differ from the reference
ordering but remain executable, whereas infeasible plans violate physical
or logical prerequisites. Table~\ref{tab:valid-alternatives} reports
Logical Progression and Temporal Alignment separately.

\begin{table}[t]
\centering
\small
\setlength{\tabcolsep}{8pt}
\begin{tabular}{@{}lcc@{}}
\toprule
\textbf{Subset} &
\shortstack{\textbf{Logical}\\\textbf{Progression}} &
\shortstack{\textbf{Temporal}\\\textbf{Alignment}} \\
\midrule
Valid alternative plans ($n=50$) & 4.04 & 1.52 \\
Infeasible plans ($n=50$) & 0.42 & 0.40 \\
\bottomrule
\end{tabular}
\caption{\textbf{Valid alternative and infeasible plans.}
Valid alternatives retain high Logical Progression despite low
Temporal Alignment, while infeasible plans score low on both
criteria. Scores are on a 0--5 scale.}
\label{tab:valid-alternatives}
\end{table}

\subsection{Effect of history length}
\label{app:history-length}

We study the effect of history length by truncating training
histories to at most 2, 5, 10, or 15 steps, using Qwen2.5-3B
with one RL run per setting. The evaluation protocol is
unchanged. Table~\ref{tab:history-length} shows that
\methodname{}-RL improves over Base even with a maximum of
two history steps. The best results on both splits are
obtained with a maximum of five steps. The default cap of 10 steps matches the average sequence length across the datasets.

\begin{table}[t]
\centering
\small
\setlength{\tabcolsep}{8pt}
\begin{tabular}{@{}lcc@{}}
\toprule
\textbf{Configuration} & \textbf{In-domain} & \textbf{Zero-shot} \\
\midrule
Base (3B) & 31.4 & 23.1 \\
\methodname{}-RL (Max = 2) & 37.3 & 38.5 \\
\methodname{}-RL (Max = 5) & \textbf{43.6} & \textbf{41.5} \\
\methodname{}-RL (Max = 10, original) & 39.2 & 39.4 \\
\methodname{}-RL (Max = 15) & 38.4 & 41.2 \\
\bottomrule
\end{tabular}
\caption{\textbf{Effect of history length (Qwen2.5-3B).}
\methodname{}-RL improves over Base even with two history
steps, with the highest scores at a maximum of five steps.
Macro rubric score (\%).}
\label{tab:history-length}
\end{table}

\subsection{Reward fidelity}
\label{app:reward-fidelity}

We examine how the grounding reward agrees with the LLM judge
at the system level and under controlled perturbations. We use
the continuous reward before the final saturation step.
Table~\ref{tab:reward-fidelity} compares the two scores across
training schemes. Both identify RL as the best method, but
the grounding reward ranks SFT above Base, unlike the judge.

\begin{table}[t]
\centering
\small
\setlength{\tabcolsep}{8pt}
\begin{tabular}{@{}lcc@{}}
\toprule
\textbf{Model} & \textbf{Grounding reward} & \textbf{Judge score} \\
\midrule
RL   & \textbf{0.503} & \textbf{0.393} \\
SFT  & 0.402 & 0.194 \\
Base & 0.385 & 0.299 \\
\bottomrule
\end{tabular}
\caption{\textbf{System-level reward fidelity.}
The grounding reward and LLM judge both rank RL first,
but disagree on the ordering of SFT and Base.}
\label{tab:reward-fidelity}
\end{table}

We also perturb ground-truth plans from Ego-Exo4D by introducing
swaps, omissions, and insertions at increasing noise levels.
Table~\ref{tab:reward-perturbation} shows that both scores
decrease monotonically with corruption. Across the five noise
levels, their Pearson and Spearman correlations are 0.927
and 1.000, respectively.

\begin{table}[t]
\centering
\small
\setlength{\tabcolsep}{8pt}
\begin{tabular}{@{}ccc@{}}
\toprule
\textbf{Noise level} & \textbf{Grounding reward} & \textbf{Judge score} \\
\midrule
10\% & 0.342 & 0.746 \\
25\% & 0.329 & 0.516 \\
50\% & 0.288 & 0.352 \\
75\% & 0.214 & 0.203 \\
90\% & 0.154 & 0.117 \\
\bottomrule
\end{tabular}
\caption{\textbf{Reward fidelity under procedural perturbations.}
Both grounding reward and judge score decrease with increasing
corruption of ground-truth plans from Ego-Exo4D.}
\label{tab:reward-perturbation}
\end{table}

\paragraph{Failure cases.}
Manual inspection also reveals disagreements between the reward
and the judge. For the goal of grilling steak, a continuation
that restarts an already ongoing procedure receives a grounding
reward of 0.761 but a judge score of 0.067. Conversely, a correct
single-step continuation for replacing a door knob receives a
grounding reward of 0.000 but a judge score of 1.000.
The first case illustrates limited sensitivity to the current
procedural state, while the second shows that the history-baseline
correction can penalize a short continuation sharing vocabulary
with the history. The grounding reward is therefore a proxy for
procedural quality rather than a direct measure of execution
correctness.

\subsection{Per-dataset breakdown of main results}
\label{app:perdataset}

Table~\ref{tab:appendix-perdataset} reports the per-dataset macro rubric score underlying the bar plot in Figure~\ref{fig:main}.

\begin{table}[h]
\centering
\caption{\textbf{Per-dataset main results: macro rubric score (\%).} Underlying numbers for Figure~\ref{fig:main}. In-domain (left of the divider) and zero-shot (right) splits are macro-averaged at the two rightmost columns. All Qwen models use the Socratic input configuration and annotated supervision. CC4D: CaptainCook4D; CrTk: CrossTask; EgPL: EgoProceL; EgPER: EgoPER; EgEx: Ego-Exo4D.}
\label{tab:appendix-perdataset}
\small
\setlength{\tabcolsep}{4pt}
\begin{tabular}{l cccc cccc cc}
\toprule
 & \multicolumn{4}{c}{\textbf{In-domain}} & \multicolumn{3}{c}{\textbf{Zero-shot}} & & \multicolumn{2}{c}{\textbf{Macro avg.}} \\
\cmidrule(lr){2-5}\cmidrule(lr){6-8}\cmidrule(l){10-11}
\textbf{Model} & CC4D & COIN & CrTk & EgPL & EgPER & EgEx & NIV & & In & ZS \\
\midrule
Qwen2.5-0.5B (Base)            &  1.9 &  1.9 &  1.7 &  0.4 &  0.9 &  0.4 &  0.9 & &  1.5 &  0.7 \\
Qwen2.5-0.5B (\methodname{}-RL)&  5.0 & 12.0 &  9.9 &  9.5 &  6.7 &  5.8 &  9.5 & &  9.1 &  7.3 \\
\midrule
Qwen2.5-3B (Base)              & 21.4 & 35.2 & 35.9 & 33.1 & 18.8 & 25.3 & 25.3 & & 31.4 & 23.1 \\
Qwen2.5-3B (\methodname{}-RL)  & 33.1 & 42.7 & 38.6 & 42.5 & 30.3 & 36.9 & 51.0 & & 39.2 & 39.4 \\
\midrule
Qwen2.5-7B (Base)              & 32.2 & 42.4 & 40.8 & 41.6 & 23.7 & 30.4 & 43.5 & & 39.2 & 32.5 \\
Qwen2.5-7B (\methodname{}-RL)  & 39.1 & 46.7 & 48.0 & 52.4 & 38.4 & 44.0 & 55.9 & & 46.6 & 46.1 \\
\bottomrule
\end{tabular}
\end{table}

\subsection{Comparison to closed-source frontier models}
\label{app:frontier}

For reference, Table~\ref{tab:appendix-frontier} reports two closed-source frontier models on the same benchmarks, included as a calibration that the rubric is not saturated rather than as a competitive comparison. Both frontier models receive Socratic-style textual histories generated by Qwen2.5-VL-3B, matching the input format of our Socratic configuration.

Two caveats apply when reading these numbers. \textbf{Capacity:} these systems operate at parameter counts roughly two orders of magnitude beyond our 7B configuration, so absolute differences reflect scale at least as much as method. \textbf{Training-data opacity:} the seven evaluation datasets we use are all publicly released and well documented, and we have no visibility into closed-source pretraining mixtures; the fact that frontier models score notably higher on our zero-shot split (NIV, EgoPER, Ego-Exo4D) than on the in-domain split, the reverse of what we observe for our Qwen runs, is consistent with possible contamination on the smaller zero-shot benchmarks but cannot be verified from the outside. We therefore treat these numbers as an upper envelope of what fully unconstrained access to capacity and data can produce, not as a target our method aims to match.

\begin{table}[h]
\centering
\caption{\textbf{Closed-source frontier models as an upper-envelope reference.} Per-dataset macro rubric score (\%) for two frontier models alongside our 7B \methodname{}-RL macro averages. See caveats in the section text; not a fair comparison. CC4D: CaptainCook4D; CrTk: CrossTask; EgPL: EgoProceL; EgPER: EgoPER; EgEx: Ego-Exo4D.}
\label{tab:appendix-frontier}
\begin{minipage}{0.72\linewidth}
\centering
\resizebox{\linewidth}{!}{%
\setlength{\tabcolsep}{3pt}
\begin{tabular}{@{}l cccc ccc cc@{}}
\toprule
 & \multicolumn{4}{c}{\textbf{In-domain}} & \multicolumn{3}{c}{\textbf{Zero-shot}} & \multicolumn{2}{c}{\textbf{Macro}} \\
\cmidrule(lr){2-5}\cmidrule(lr){6-8}\cmidrule(l){9-10}
\textbf{Model} & CC4D & COIN & CrTk & EgPL & EgPER & EgEx & NIV & In & ZS \\
\midrule
Gemini-3.1-Flash  & 59.8 & 43.5 & 54.8 & 63.7 & 76.3 & 68.2 & 62.6 & 55.5 & 69.0 \\
GPT-5.4-mini      & 62.0 & 44.8 & 56.3 & 62.1 & 76.2 & 71.2 & 61.6 & 56.3 & 69.7 \\
\bottomrule
\end{tabular}%
}
\end{minipage}\hfill
\begin{minipage}{0.26\linewidth}
\centering
\resizebox{\linewidth}{!}{%
\setlength{\tabcolsep}{4pt}
\begin{tabular}{@{}l cc@{}}
\toprule
\textbf{Reference} & In & ZS \\
\midrule
\methodname{}-RL 7B & 46.6 & 46.1 \\
Qwen2.5-7B (Base)   & 39.2 & 32.5 \\
\bottomrule
\end{tabular}%
}
\end{minipage}
\end{table}

\subsection{Qualitative results}
\label{app:qualitatives}

Table~\ref{tab:all_qualitatives} compares the predicted continuations of the Base model, SFT, and \methodname{}-RL against the reference for six representative test prompts spanning cooking, crafting, and bicycle maintenance tasks. Examples are organized to span the spectrum of outcomes: two clear wins for \methodname{} (Examples~1--2), one tie at parity with Base (Example~3), and three failure modes (Examples~4--6). Each entry shows the goal, the history of steps already executed, and the predicted continuation; the rightmost column is the rubric score (sum of the six $0$--$5$ criteria from Appendix~\ref{app:rubric}, max $30$).

\paragraph{Where \methodname{} helps.} \methodname{}-RL exhibits stronger procedural reasoning when the task requires building on the history rather than restarting it. In Example~1, it correctly progresses to adding toppings and seasoning to an already-plated salad, avoiding the redundant preparation steps the Base model invents. In Example~2 it uniquely recognizes the need to repeat earlier steps to construct the flower crown, achieving the highest score (24/30). Example~3 is a competitive tie: both Base and \methodname{} score 9/30 by correctly applying lubricant and wiping off the excess; SFT misses the wiping step.

\paragraph{Failure modes.} The framework is not without limitations. Example~5 shows a spatial failure: \methodname{} regresses to mixing ingredients in a bowl despite the visual history clearly showing they are already plated. Example~6 shows a more severe causal failure: the model proposes reversing the procedure (removing the wheel) rather than completing the installation. These cases highlight that the grounding reward, while effective at improving plan plausibility on average, does not guarantee consistency with fine-grained spatial cues or with the directionality implied by the goal; closing these gaps remains future work.

\begingroup
\small
\renewcommand{\arraystretch}{1.3}
\begin{longtable}{p{4cm} l p{6.5cm} c}
\caption{\textbf{Qualitative results.} Continuations predicted by the Base model, SFT, and \methodname{}-RL for six representative prompts, alongside the reference and the rubric score (max $30$). Examples~1--2 are wins for \methodname{}, Example~3 is a tie, Examples~4--6 are failure modes.}
\label{tab:all_qualitatives} \\

\toprule
\textbf{Context / Goal} & \textbf{Model} & \textbf{Generated Plan} & \textbf{Score} \\
\midrule
\endfirsthead

\multicolumn{4}{c}%
{{\bfseries Table \thetable\ continued from previous page}} \\
\toprule
\textbf{Context / Goal} & \textbf{Model} & \textbf{Generated Plan} & \textbf{Score} \\
\midrule
\endhead

\midrule
\multicolumn{4}{r}{{Continued on next page}} \\
\endfoot

\bottomrule
\endlastfoot

\textbf{Example 1} \newline \newline 
\textbf{Goal:} Prepare tomato mozzarella salad \newline \newline 
\textbf{History:} \newline 
1. Washes tomato under running water. \newline 
2. Chops tomato into slices. \newline 
3. Arranges tomato slices on a plate. \newline 
4. Sprinkles seasoning over the tomato slices. \vspace{0.3cm} \newline 
\noindent \fbox{
    \includegraphics[width=3.5cm,height=3.5cm,keepaspectratio]{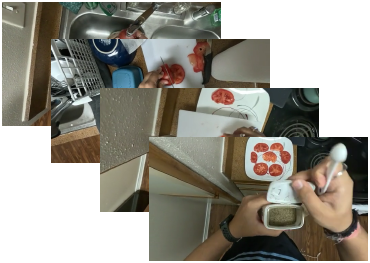}
}
& 
\multicolumn{3}{@{}l@{}}{
    \begin{tabular}[t]{@{} l p{5.5cm} c @{}}
    Ground Truth & Finish the tomato mozzarella salad by sprinkling it with Italian seasoning. / Distribute shredded mozzarella cheese evenly across the tomatoes on the platter. / Sprinkle salt over the tomato slices. / Drizzle extra-virgin olive oil, approximately 1 tablespoon, evenly across the platter. & - \\ \cmidrule{1-3}
    Base & 5. Washes and dries the mozzarella cheese. / 6. Slices the mozzarella cheese into thin rounds. / 7. Places the mozzarella cheese slices on top of the tomato slices. / 8. Drizzles olive oil and balsamic vinegar over the tomato and mozzarella salad, and adds fresh basil leaves for garnish. & 14/30 \\ \cmidrule{1-3}
    SFT & 5. Sprinkle seasonings evenly across the tomato slices. / 6. Combine mozzarella cheese with tomato. / 7. Fold the salad ingredients together. & 13/30 \\ \cmidrule{1-3}
    \textbf{RECIPE} & 5. Sprinkles mozzarella cheese over the tomato slices. / 6. Drizzles olive oil over the cheese. / 7. Sprinkles salt and pepper over the dish. / 8. Places the plate in the refrigerator for 10 minutes to let the flavors meld together. & \textbf{23/30}
    \end{tabular}
} \\ 
\midrule

\textbf{Example 2} \newline \newline 
\textbf{Goal:} Craft flower crown \newline \newline 
\textbf{History:} \newline 
1. Tie a loop with twine using a pink stapler. \newline 
2. Cut a branch with scissors. \newline 
3. Attach the branch to the loop. \vspace{0.3cm} \newline 
\noindent \fbox{
\includegraphics[width=3.5cm,height=3.5cm,keepaspectratio]{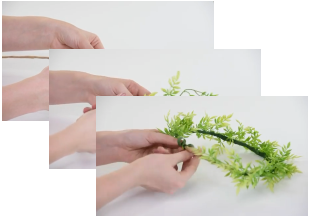}
}
& 
\multicolumn{3}{@{}l@{}}{
    \begin{tabular}[t]{@{} l p{5.5cm} c @{}}
    Ground Truth & Gather and arrange flowers. / Secure the flower to the crown frame. & - \\ \cmidrule{1-3}
    Base & 4. Gather a variety of fresh flowers and greenery to create the flower crown. / 5. Arrange the flowers and greenery on the branch, securing them with floral wire or additional twine to create a cohesive design. & 22/30 \\ \cmidrule{1-3}
    SFT & 4. Arrange flowers into a circular shape. & 17/30 \\ \cmidrule{1-3}
    \textbf{RECIPE} & 4. Repeat steps 2 and 3 to add more branches to the loop, creating a flower crown shape. / 5. Add flowers to the crown by attaching them to the branches using floral wire or ribbon. & \textbf{24/30}
    \end{tabular}
} \\ 
\midrule

\textbf{Example 3} \newline \newline 
\textbf{Goal:} Clean and Lubricate Chain \newline \newline 
\textbf{History:} \newline 
1. The person holds a bottle of lubricant in one hand and a cloth in the other hand. \vspace{0.3cm} \newline 
\noindent \fbox{
\includegraphics[width=3.5cm,height=3.5cm,keepaspectratio]{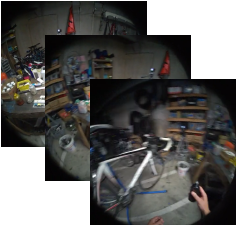}
}
& 
\multicolumn{3}{@{}l@{}}{
    \begin{tabular}[t]{@{} l p{5.5cm} c @{}}
    Ground Truth & Obtain the necessary cleaning equipment. / Wipe the chain with a cloth to remove any excess moisture. & - \\ \cmidrule{1-3}
    \textbf{Base} & 1. The person opens the bottle of lubricant and pours some of the lubricant onto the chain. / 2. The person uses the cloth to wipe off any excess lubricant from the chain, ensuring it is evenly distributed. & \textbf{9/30} \\ \cmidrule{1-3}
    SFT & 2. The person pours lubricant onto the chain and then applies it to the chain. & 4/30 \\ \cmidrule{1-3}
    \textbf{RECIPE} & 2. The person sprays lubricant onto the chain, ensuring even coverage. / 3. The person uses the cloth to wipe away any excess lubricant, leaving the chain properly coated and clean. & \textbf{9/30}
    \end{tabular}
} \\ 
\midrule

\textbf{Example 4} \newline \newline 
\textbf{Goal:} Fix a Flat Tire \newline \newline 
\textbf{History:} \newline 
1. The person removes the wheel from the bike. \vspace{0.3cm} \newline 
\noindent \fbox{
\includegraphics[width=3.5cm,height=3.5cm,keepaspectratio]{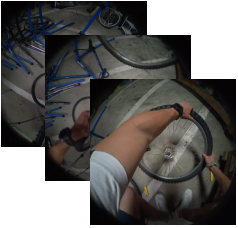}
}
& 
\multicolumn{3}{@{}l@{}}{
    \begin{tabular}[t]{@{} l p{5.5cm} c @{}}
    Ground Truth & Obtain a replacement inner tube. / Mount the tube onto the wheel. / Mount the tire onto the wheel. / Completely fill the tire with air. & - \\ \cmidrule{1-3}
    \textbf{Base} & 2. The person removes the tire and then the tube. / 3. The person replaces the tube. / 4. The person replaces the tire and then inflates the tube. & \textbf{15/30} \\ \cmidrule{1-3}
    SFT & 1. Inflate the new tube to the recommended pressure level. / 2. Remove the old tube from the wheel and install the new tube inside the tire. / 3. Reinstall the wheel onto the bike frame. & 10/30 \\ \cmidrule{1-3}
    RECIPE & 2. Locate the puncture in the tire and remove the valve stem cap. / 3. Carefully remove the inner tube from the tire, making sure not to pinch any rubber. / 4. Insert the new inner tube into the tire, making sure the valve stem is aligned correctly. / 5. Reinstall the wheel onto the bike, ensuring that the spokes are evenly tensioned. / 4. Inflate the tire to the recommended pressure level and check for any leaks. & 10/30
    \end{tabular}
} \\ 
\midrule

\textbf{Example 5} \newline \newline 
\textbf{Goal:} Prepare tomato mozzarella salad \newline \newline 
\textbf{History:} \newline 
1. Rinse tomato under running water. \newline 
2. Place tomato on cutting board. \newline 
3. Slice tomato into wedges. \newline 
4. Transfer tomato slices to plate. \vspace{0.3cm} \newline 
\noindent \fbox{
\includegraphics[width=3.5cm,height=3.5cm,keepaspectratio]{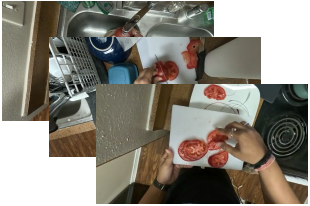}
}
& 
\multicolumn{3}{@{}l@{}}{
    \begin{tabular}[t]{@{} l p{5.5cm} c @{}}
    Ground Truth & Sprinkle the platter with 1/4 teaspoon of black pepper. / Finish the tomato mozzarella salad by sprinkling it with Italian seasoning. / Scatter shredded mozzarella cheese evenly across the tomatoes on the platter. / Sprinkle salt evenly over the tomato slices. / Drizzle approximately one tablespoon of extra-virgin olive oil evenly over the entire platter. & - \\ \cmidrule{1-3}
    Base & 1. Wash mozzarella cheese under running water. / 2. Place mozzarella cheese on cutting board. / 3. Slice mozzarella cheese into thin slices. / 4. Transfer mozzarella slices to the plate with tomato slices. / 5. Mix the tomato and mozzarella slices together. & 7/30 \\ \cmidrule{1-3}
    \textbf{SFT} & 5. Sprinkle seasonings over tomato mozzarella salad. / 6. Combine mozzarella cheese and tomatoes in a bowl. / 7. Fold the salad ingredients together. / 8. Combine mozzarella cheese with tomatoes in a bowl. / 9. Drizzle a vinaigrette over the salad. & \textbf{11/30} \\ \cmidrule{1-3}
    RECIPE & 5. Add a small amount of olive oil to a mixing bowl. / 6. Add a handful of fresh mozzarella cheese to the bowl. / 7. Add a pinch of salt and pepper to the bowl. / 8. Use a fork to gently mix the ingredients together until the cheese is evenly coated with oil. / 9. Arrange the tomato slices on a serving dish. / 10. Sprinkle the cheese mixture over the tomato slices. / 11. Serve immediately and enjoy your tomato mozzarella salad! & 9/30
    \end{tabular}
} \\ 
\midrule

\textbf{Example 6} \newline \newline 
\textbf{Goal:} Install a Wheel \newline \newline 
\textbf{History:} \newline 
1. The person holds a wrench and tightens a bolt on the bicycle wheel. \vspace{0.3cm} \newline 
\noindent \fbox{
\includegraphics[width=3.5cm,height=3.5cm,keepaspectratio]{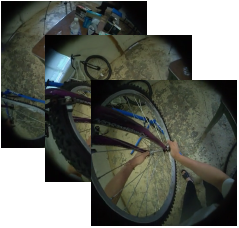}
}
& 
\multicolumn{3}{@{}l@{}}{
    \begin{tabular}[t]{@{} l p{5.5cm} c @{}}
    Ground Truth & Securing the Wheel in Place. & - \\ \cmidrule{1-3}
    Base & 2. The person removes the tire from the bicycle wheel. / 3. The person installs the new wheel onto the bicycle frame. & 10/30 \\ \cmidrule{1-3}
    \textbf{SFT} & 2. The person attaches the wheel to the bicycle frame. & \textbf{24/30} \\ \cmidrule{1-3}
    RECIPE & 2. Remove the wheel from the bicycle frame. / 3. Apply grease to the bearings and reassemble the wheel. & 2/30
    \end{tabular}
} \\ 

\end{longtable}
\endgroup

\section{Limitations}
\label{app:limitations}
\methodname{} is designed to learn from broader procedural evidence than single annotated trajectories, but its evaluation is still constrained by the benchmarks available today. Most datasets provide one reference continuation per example, so even a flexible LLM-as-judge protocol cannot fully capture the space of valid alternative plans, making our reported numbers a conservative estimate of open-ended planning quality. The reward also reflects the coverage of the verification corpus: HowToCaption/HowTo100M provides large-scale procedural evidence across many domains, but its distribution is not exhaustive, and rare procedures, unusual tool choices, or underrepresented linguistic and cultural conventions may be less strongly rewarded than common ones. Finally, our verifier is based on textual procedural alignment: it improves trajectory-level plausibility but does not explicitly check fine-grained visual state. Multi-reference evaluation and multimodal verifiers are promising directions for future work.

\section{Societal Impact}
\label{app:societal_impact}
\methodname{} may contribute to procedural assistants that guide users through everyday tasks such as cooking, repair, and other step-by-step activities. While such systems could improve accessibility and support non-expert users, they also introduce risks when generated plans are incorrect, incomplete, or unsafe, especially for tasks involving tools, heat, electricity, allergens, or other hazards. Since our reward is computed from large-scale instructional-video captions, the model may inherit geographical, cultural, linguistic, or demographic biases present in the source corpus, such as particular ways of preparing food or performing household tasks. These biases may limit the quality of assistance for users from underrepresented contexts. We therefore view \methodname{} as a step toward assistive procedural guidance rather than autonomous instruction in safety-critical settings; deployment should include human oversight, task-specific safety constraints, and evaluation on diverse users and domains.

\end{document}